\documentclass[10pt,twocolumn,letterpaper]{article}
 \usepackage{cvpr}
\usepackage{amsfonts}
\usepackage{amsmath}
\usepackage{bm}
\usepackage{mleftright}
\usepackage{multirow}
\usepackage{pgfplots}
\usepackage{siunitx}
\usepackage{xcolor}
\usepackage{xspace}

\definecolor{cvprblue}{rgb}{0.21,0.49,0.74}
\usepackage[pagebackref,breaklinks,colorlinks,allcolors=cvprblue]{hyperref}
\usetikzlibrary{tikzmark,positioning}

\pgfplotsset{compat=1.17}

\newcommand{\method}{VFMF\xspace}

\def\vv{{\boldsymbol{v}}}
\def\vz{{\boldsymbol{z}}}
\def\vf{{\boldsymbol{f}}}

\def\vmu{{\boldsymbol{\mu}}}
\def\vsigma{{\boldsymbol{\sigma}}}
\def\Enc{{\boldsymbol{\phi}}}
\def\Dec{{\boldsymbol{\psi}}}

\newcommand*{\Encoder}[1]{\textsc{Encoder}\mleft({#1}\mright)}

\definecolor{GeomColor}{RGB}{255, 0, 0}   
\definecolor{RGBArtColor}{RGB}{0, 47, 255} 
\definecolor{ShapeColor}{RGB}{0, 255, 9}   

\newcommand{\geomdist}{\textcolor{GeomColor}{geometric distortions}}
\newcommand{\rgbart}{\textcolor{RGBArtColor}{RGB-space artefacts}}
\newcommand{\shapeinc}{\textcolor{ShapeColor}{shape inconsistencies}}

\makeatletter
\renewcommand{\paragraph}{%
  \@startsection{paragraph}{4}%
  {\z@}{-0.5em}{-0.5em}%
  {\normalfont\normalsize\bfseries}%
}
\makeatother

\def\paperID{3242}
\def\confName{CVPR}
\def\confYear{2026}

\title{\method: World Modeling by \underline{F}orecasting \underline{V}ision \underline{F}oundation \underline{M}odel Features}

\author{
\begin{tabular}{cccc}
Gabrijel Boduljak &
Yushi Lan &
Christian Rupprecht &
Andrea Vedaldi \\
\small VGG, University of Oxford &
\small VGG, University of Oxford &
\small VGG, University of Oxford &
\small VGG, University of Oxford \\
\small \texttt{gabrijel@robots.ox.ac.uk} &
&
&
\\
\end{tabular}
}

\begin{document}
\twocolumn[
\maketitle
\begin{center}
\makebox(500,200){%
    \includegraphics[scale=0.2]{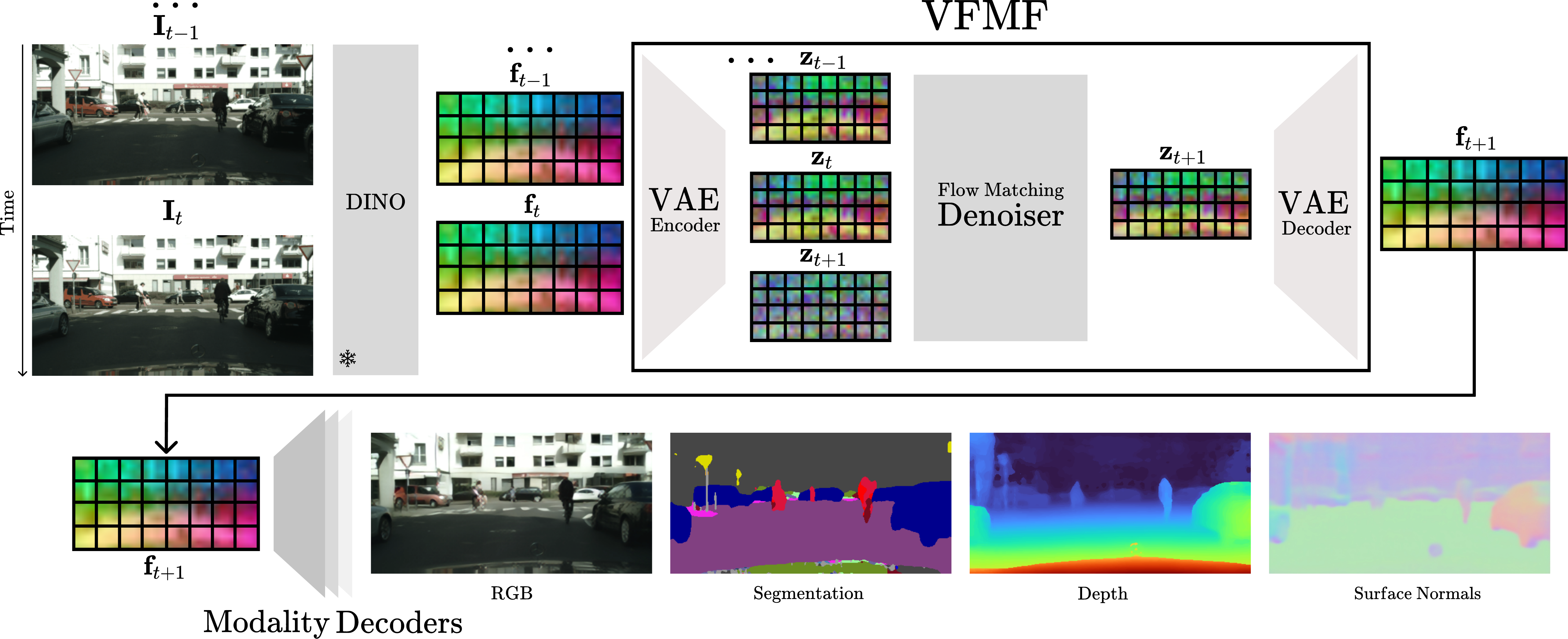}
}
\end{center}
\captionof{figure}{ \textbf{An overview of our method.} Our world-modeling method, \textbf{\method}, \textit{autoregressively generates} diverse futures in the \textit{latent space} of a \textit{foundation model}, translatable into downstream modalities, such as segmentation, depth, surface normals and even RGB.}
\label{fig:teaser}
\vspace{1em}
]


\begin{abstract}
Forecasting from partial observations is central to world modeling.
Many recent methods represent the world through images, and reduce forecasting to stochastic video generation. Although such methods excel at realism and visual fidelity, predicting pixels is computationally intensive and not directly useful in many applications, as it requires translating RGB into signals useful for decision making.
An alternative approach uses features from vision foundation models (VFMs) as world representations, performing deterministic regression to predict future world states. These features can be directly translated into actionable signals such as semantic segmentation and depth, while remaining computationally efficient. However, deterministic regression averages over multiple plausible futures, undermining forecast accuracy by failing to capture uncertainty.
To address this crucial limitation, we introduce a generative forecaster that performs autoregressive flow matching in VFM feature space. Our key insight is that generative modeling in this space requires encoding VFM features into a compact latent space suitable for diffusion. We show that this latent space preserves information more effectively than previously used PCA-based alternatives, both for forecasting and other applications, such as image generation.
Our latent predictions can be easily decoded into multiple useful and interpretable output modalities: semantic segmentation, depth, surface normals, and even RGB\@. With matched architecture and compute, our method produces sharper and more accurate predictions than regression across all modalities.
Our results suggest that stochastic conditional generation of VFM features offers a promising and scalable foundation for future world models.

\href{https://vfmf.gabrijel-boduljak.com/}{\underline{\textcolor{cyan}{Project Page}}}
\;\textbar\;
\href{https://github.com/gboduljak/vfmf}{\underline{\textcolor{cyan}{GitHub}}}

\end{abstract}
\section{Introduction}%
\label{sec:intro}

One of the key challenges in world modeling is to forecast future states of a scene from partial observations of its past~\cite{ha18world}.
There is significant debate on how to tackle this problem, starting with the choice of world representation.
Many have suggested that world models should be based on video generators~\cite{li24sora,nvidia25cosmos,bruce2024genie,parker-holder24genie}.
These are appealing for three reasons.
First, when implemented incrementally or autoregressively, they effectively predict future pixels from past ones.
Second, because they process pixels, they can be trained on enormous amounts of video data with minimal curation and manual supervision.
Third, the quality of these models has improved dramatically in recent years.

A key downside of pixel-based models, however, is that predicting pixels is neither necessary nor directly useful in many applications.
Predicting pixels is computationally intensive, and generators often make physically implausible predictions~\cite{kang2024how,boduljak2025happensnextanticipatingfuture,motamed2025generativevideomodelsunderstand}.
The output of world models should be actionable and serve as the basis for decision-making.
For example, we may decide to steer a car if failing to do so is predicted to cause an accident.
However, generating an \emph{image} of an accident is not the same as understanding that an accident may occur.
The latter requires parsing the image to extract its meaning.
Thus, by generating pixels, we may be doing unnecessary work and producing outputs that are needlessly complex to interpret.

Even before video generators became popular, authors considered alternative, more useful, and compact representations for scene forecasting.
A particularly interesting approach is based on the representations computed by vision foundation models (VFMs), such as DINO-based models~\cite{caron21emerging,oquab24dinov2:,simeoni25dinov3}.
These spaces tend to capture information that is more easily decoded into semantic properties (\eg, object classes) that are directly actionable.
In fact, they are semi-dense and can be decoded into geometric properties such as depth maps, too, which may also be more directly useful to an agent operating in the physical world.
Furthermore, they may disregard low-level image details that are irrelevant to the task at hand, whose modeling may be wasteful.

Because of these observations, several authors~\cite{dino-foresight,dino-wm,dino-world} have suggested basing world forecasting on these feature spaces.
Once a prediction is made in the VFM feature space, lightweight decoders can extract semantic and geometric properties, or even reconstruct RGB images if needed.
However, unlike video generators, which are inherently stochastic and can model uncertainty about the future, these models have so far focused on \emph{deterministic prediction}.
We argue that deterministic forecasting is fundamentally ill-suited for world models, as the state of the world is almost always only partially observed.
As is well known, and as we further show in our experiments, deterministic prediction of ambiguous targets yields blurry predictions that compound during rollout (\cref{fig:regressionblurry}).
This is particularly evident when context length is variable or short, as is common in practical deployments.
This motivates the following question: \emph{How can we build stochastic forecasters in the feature spaces of foundation models, and what are the benefits of doing so?}

To answer this question, we start by noting that even video generators operate in latent spaces (\eg, via latent diffusion); the difference is that these are tailored to represent pixels.
This suggests that similar technology should be applicable to forecasting in VFM feature spaces.
Our first contribution is thus to replace deterministic regression in models like~\cite{dino-foresight} with \emph{stochastic conditional generation} via diffusion-style models~\cite{lipman2023flowmatching,liu2022rectifiedflow}.

Our second contribution is to show how to perform this diffusion effectively.
Effective diffusion latents are typically low-dimensional and trained with specialized variational autoencoders (VAEs)~\cite{Kingma2013AutoEncodingVB}, whereas features extracted by VFMs are often high-dimensional, making diffusion ill-conditioned and unstable~\cite{Hoogeboom_2025_CVPR}.
One solution is principal component analysis (PCA)–based compression (\eg, ReDi~\cite{ReDi}), and this has been used in deterministic VFM forecasting~\cite{dino-foresight}.
We argue that this is not very effective because it discards too much information (\cref{fig:vae_vs_pca}).
Instead, we propose learning a \emph{compact latent space on top of VFM features} using a new VAE for this purpose.
We show that our VAE-compressed VFM feature space yields a compact, well-conditioned latent space that better preserves semantic and geometric information.
Within a latent diffusion model, it enables uncertainty-aware, temporally coherent forecasting that remains robust across different context lengths.
The forecasts can then be decoded to output modalities such as semantic segmentation, depth, and normals, as well as RGB, with lightweight decoders, without the need to interpret pixel-space predictions.

Besides forecasting, we demonstrate the advantages of this VAE over PCA in other applications, such as image generation~\cite{ReDi}. This is promising, given the numerous works that apply PCA to VFMs, such as DINO~\cite{oquab24dinov2:,simeoni25dinov3}.

Compared to regression-based forecasters, our approach produces sharper and more accurate semantic, depth, and normal predictions, particularly when context is short.
This holds even after carefully calibrating architecture and compute budgets to be comparable with the regression baselines, indicating that the key differences are the ability to model uncertainty and the choice of latent space.

To summarize, our contributions are as follows:
(i) We show that deterministic, regression-based forecasters perform poorly with variable and short context: they collapse multimodal\footnote{In a distributional sense.} predictions to means that are not necessarily meaningful.
(ii) We use \emph{autoregressive flow matching} for forecasting in VFM feature spaces, yielding uncertainty-aware, coherent predictions that work well with different context lengths.
(iii) We introduce a \emph{VAE-compressed VFM feature space} that preserves useful information much better than the PCA compression used in prior work.
(iv) We demonstrate significant improvements in the forecasting of semantic and geometric quantities, as well as RGB.
\section{Related Work}%
\label{sec:related-work}

\paragraph{Vision Foundation Models.}

Vision foundation models (VFMs) have transformed visual representation learning by training on large-scale image, video, and multimodal datasets. 
Self-supervised approaches, such as DINO and DINOv2~\cite{dino,oquab2024dinov,simeoni2025dinov3}, learn rich visual embeddings through self-distillation. In contrast, vision-language models like CLIP~\cite{radford2021learning,fang2024eva} align visual and textual spaces, while SAM~\cite{kirillov2023segment} enables open-set segmentation.  
Video and 3D extensions, including VideoMAE~\cite{assran2025vjepa2,carreira2024scaling} and VGGT-like models~\cite{wang2025vggt}, extend these representations across space and time.  
Although primarily designed for perception, recent work shows that pretrained VFM features can also benefit generation and reconstruction tasks~\cite{RAE,ReDi}.  
Building on this insight, we explore how the latent spaces of VFMs can serve as compact, semantically meaningful substrates for generative world modeling.  
Our proposed VAE-based feature compression preserves substantially more information than PCA-based alternatives~\cite{ReDi}, enabling faithful multimodal decoding.

\paragraph{World Modeling in Feature Space.}

World models~\cite{ha_schmidhuber_nips18} aim to predict the temporal evolution of an environment from past observations, traditionally in pixel or state space.  
Recent works such as DINO-Foresight~\cite{dino-foresight}, DINO-WM~\cite{dino-wm}, and DINO-World~\cite{dino-world} reformulate this task as \emph{future feature prediction} in the latent space of pretrained VFMs, following a ``back to the features'' philosophy~\cite{dino-world}.  
By operating in semantically structured latent spaces, these methods achieve more stable and interpretable forecasting than pixel-based alternatives, enabling downstream tasks such as semantic and geometric prediction~\cite{luc2018predictingfutureinstancesegmentation,dino-foresight}.  
However, these models are fundamentally regression-based and deterministic, which leads to \emph{averaged} and unrealistic predictions under multimodal or uncertain futures.  
Our work extends this line by introducing a generative, stochastic formulation that explicitly models uncertainty via conditional distributions in the VFM latent space.

\paragraph{Generative World Modeling.}

Video generators are considered by many as proxies to world models.  
Diffusion- and transformer-based approaches such as Sora~\cite{sora}, Genie~\cite{bruce2024genie}, and Cosmos~\cite{nvidia2025cosmosreason1physicalcommonsense} synthesize realistic video continuations conditioned on text or context frames.  
Domain-specific systems like GAIA~\cite{hu2023gaia} and VAVAM~\cite{vavam2025} focus on driving or embodied navigation~\cite{bar2024navigation}.  
While these pixel-space models achieve impressive visual fidelity, they require heavy computation, struggle with generalization across domains, and often lack physical consistency~\cite{kang2024how}.  
In contrast, our approach models dynamics directly in the VAE-compressed latent space of VFMs, capturing high-level semantics and physical structure while remaining computationally efficient.  
Specifically, we adopt \emph{flow matching}~\cite{lipman2023flowmatching,liu2022rectifiedflow} to train our generative world model, learning stochastic dynamics in a compact and semantically meaningful feature space.  
This latent generative formulation bridges the gap between regression-based feature forecasting and pixel-space video generation, and further enables a variety of downstream tasks, including segmentation, depth, normal, and RGB prediction, that prior deterministic approaches cannot achieve.

\section{Method}%
\label{sec:method}

\newcommand{\real}{\mathbb{R}}

Motivated by the fact that forecasting the future is a key part of \emph{world models}~\cite{ha_schmidhuber_nips18}, we design a model that can forecast future states of a scene represented in the latent space of vision foundation models (VFMs).
Following prior works~\cite{dino-foresight,dino-wm,dino-world}, we predict future VFM features conditioned on a \emph{variable-length} context of past observations---a realistic setting for agents where a shorter context implies higher \emph{predictive uncertainty} (more plausible futures), and thus challenges purely deterministic predictors.

In \cref{sec:regression-based-world-model} we summarize the conventional deterministic formulation and explain why it blurs under variable context.
\Cref{sec:our-autoregresive-world-model} introduces our stochastic, autoregressive formulation.
\Cref{sec:vae} presents a VAE-compressed feature space, and \cref{sec:flow} details training/inference via flow matching.
Downstream decoders are in \cref{sec:downstream}.

\subsection{Background: Deterministic VFM forecasting}%
\label{sec:regression-based-world-model}

We begin by describing the idea of deterministic forecasting in VFM feature spaces introduced by~\cite{dino-foresight,dino-wm,dino-world}.
Let a video be a sequence
\(
   \{ (\vv_t, t) \}_{t=1}^T
\)
where the tensor
\(
   \vv_t \in \real^{H'\times W'\times 3}
\)
is a video frame.
A VFM encoder (\eg, DINOv2~\cite{oquab24dinov2:}) maps each frame to features
\(
   \vf_t=\Encoder{\vv_t}\in\real^{H\times W\times D}
\).
Given a context \(\vf_{1:T}\) of length $T$, regression-based methods predict future features \(\vf_{t'}\) for \(t'>T\) and, if trained to minimize the \(\ell_2\) loss, approximate the conditional mean \(\mathbb{E}[\vf_{t'}\mid \vf_{1:t}]\).
In general, but in particular when $T$ is small (\eg, 1--2 frames), the state of the scene is underdetermined since \emph{many futures are plausible}.
A single point estimate averages these hypotheses, yielding over-smoothed predictions that become worse as the context shortens and rollouts become longer (\cref{fig:regressionblurry}).

\subsection{Stochastic VFM forecasting}%
\label{sec:our-autoregresive-world-model}

In order to address the limitations of deterministic forecasting, we propose a \emph{stochastic} model that captures uncertainty over future features.
Formally, this amounts to learning a conditional distribution \(p(\vf_{T+1}\mid \vf_{1:T})\) from which several plausible futures can be sampled.
As the context ($\mathcal{C}$) length $T$ grows, predictive uncertainty naturally decreases and forecasts sharpen; with shorter context, sample variability reflects the increasing ambiguity rather than collapsing to an average.
Conceptually, this mirrors autoregressive video generation, but we generate \emph{VFM features} rather than pixels\footnote{Despite the intent of VAEs to learn abstract representations, the latents produced by state-of-the-art RGB VAEs remain relatively low-level: they inherit the spatial grid structure of the input, closely resembling a downsampled version of the input image or video ~\cite{dieleman2025latents}.}.
We implement this with \emph{latent flow matching}~\cite{lipman2023flowmatching} on a compact feature latent (\cref{sec:vae}).

\subsubsection{Auto-encoding VFM Features}%
\label{sec:vae}

A challenge of forecasting in VFM feature space is that these features are high-dimensional, with $D$ in the hundreds or thousands of feature channels.
Prior works have suggested to PCA-compress these features~\cite{dino-foresight,ReDi}, but this discards too much information, harming generation quality.

We thus propose to train a VAE~\cite{kingma13auto-encoding} over the VFM features to obtain a generative-friendly compact latent code
\(
\vz \in \mathbb{R}^{H\times W\times D/r}
\).
The goal of the VAE is to reduce the feature dimension by a large factor $r$ (whereas the spatial dimension is preserved as the VFM features are already spatially downsampled compared to the input images).
Based on the VAE formulation, the encoder outputs a diagonal Gaussian posterior with parameters
\(
(\vmu_{\vz},\vsigma_{\vz}) = \Enc(\vf)
\),
and the decoder reconstructs
\(
\hat{\vf}=\Dec(\vz)
\)
from a sampled latent
\(
\vz \sim \mathcal{N}_{(\vmu_{\vz},\vsigma_{\vz})}.
\)
We train by optimizing the \( \beta \)-VAE loss
\[
\mathcal{L}_{\beta\text{-VAE}}
=
\mathbb{E}_{\vz\sim\mathcal{N}_{\Enc(\vf)}}
[
    \tfrac{1}{2}\|\vf-\Dec(\vz)\|_2^2
]
+ \beta \cdot
\mathbb{D}_{\mathrm{KL}}(\mathcal{N}_{\Enc(\vf)}\|\mathcal{N}_0)
\]
where \(\mathcal{N}_0\) is standard normal.
Note that frames are auto-encoded independently.

\subsection{Forecasting using Rectified Flow}%
\label{sec:flow}

Having mapped the VFM features \(\vf_t\) to compact latents \(\vz_t\) with the VAE, we now learn a stochastic forecasting model in this latent space.
Namely, instead of learning the distribution \(p(\vf_{T+1}\mid \vf_{1:T})\), we learn \(p(\vz_{T+1}\mid \vz_{1:T})\).

We do so by using \emph{rectified flow/flow matching}~\cite{lipman22flow,liu23flow}.
A velocity network
\(
\hat{\vv}_\theta(\vz^{(t)},\mathcal{C},t)
\)
is trained with the standard objective.
During training, we \emph{randomize} the length of the context ($|\mathcal{C}|=T$) in the range $1$ to $K$, so the model calibrates uncertainty to the available history, for example, learning that shorter contexts result in increased ambiguity.

At test time we sample
\(
\vz^{(0)} \sim \mathcal{N}(0,I)
\)
and integrate the learned ODE
\(
\dot{\vz}^{(t)} = \hat{\vv}_\theta(\vz^{(t)},\mathcal{C},t)
\)
from $t=0$ to $1$ to obtain
\(
\hat{\vz}_{T+1}=\vz^{(1)}
\).
We roll out autoregressively with a sliding window of length \(K\), \ie,
\(
p(\vz_{T+1}\mid \vz_{1:T})
\approx
p(\vz_{T+1}\mid \vz_{T-K+1:T})
\).

\subsection{Decoding Multiple Modalities}%
\label{sec:downstream}

Forecasting in VFM feature space simplifies decoding the prediction to several useful interpretable modalities.
For semantic segmentation, depth, and surface normals, we follow DINO-Foresight~\cite{dino-foresight} and use simple regression heads.
For RGB reconstruction, we use a ViT-B~\cite{dosovitskiy2020image} backbone with a DPT-based decoder~\cite{ranftl2021vision}, trained with LPIPS~\cite{zhang2018perceptual} and \(\ell_1\).

\paragraph{Discussion.}

\begin{figure*}
    \centering
    \includegraphics[width=0.9\linewidth]{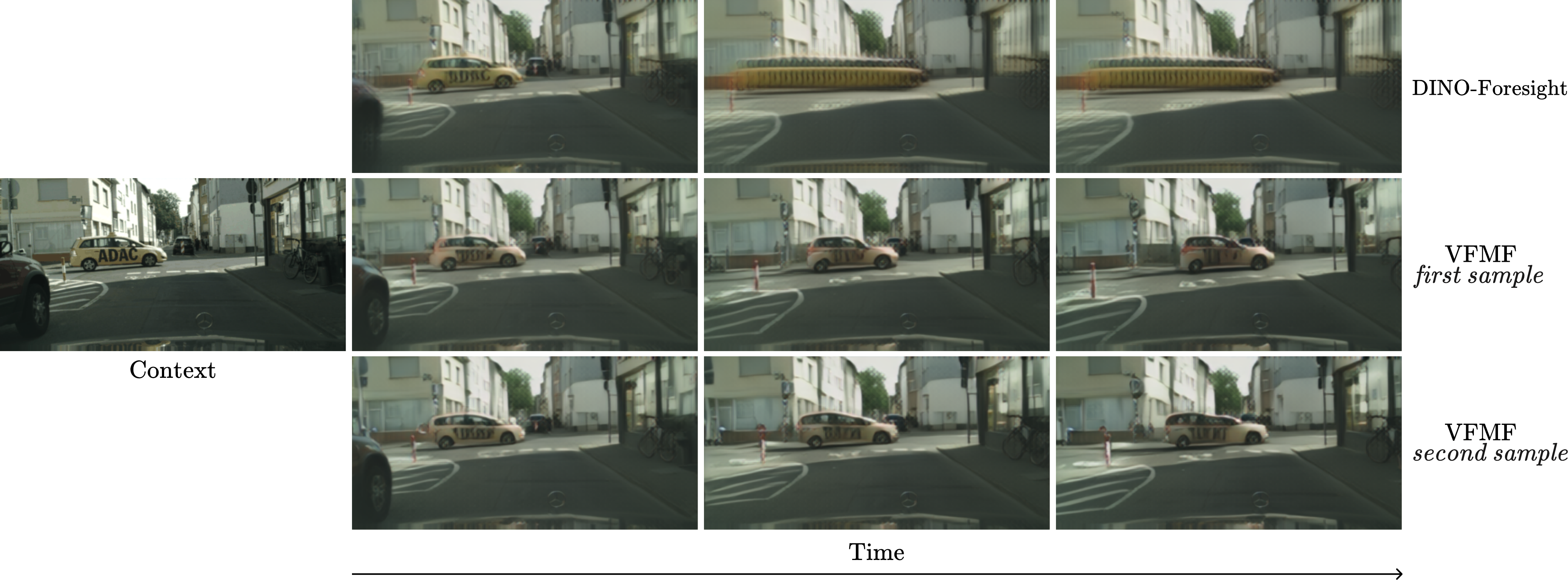}
    \caption{\textbf{Qualitative comparison of future predictions} translated into RGB domain. DINO-Foresight, a \textit{regression} baseline, is unable to model uncertainty in the motion of both the ego-vehicle and the car in the middle of the street, effectively averages all possible futures, producing blurry and physically implausible predictions. In contrast, our \textit{generative} method generates plausible futures, accurately capturing the uncertainty in unknown velocities and accelerations, that can be translated into sharp RGB or other modalities (\cref{fig:modalities}).}
    \label{fig:regressionblurry}
\end{figure*}

While it is difficult to assess the benefits of a VAE directly in VFM space, we can instead measure its effect on downstream decoding tasks.
Here we discuss one such example and investigate others in \cref{sec:experiments}.
It is well known that many high-bandwidth visual features are approximately \emph{invertible}, in the sense that the input image can be reconstructed from them~\cite{mahendran15understanding,nguyen16synthesizing}.
We therefore train an \emph{inverter} (\ie, a network that maps features back to an image) and evaluate whether inversion remains possible after compressing and decompressing the features.
\Cref{fig:regressionblurry,fig:vae_vs_pca} shows that VAE compression substantially outperforms PCA, which yields blurry reconstructions and information loss.
\section{Experiments}%
\label{sec:experiments}

In \cref{subsec:forecasting}, we begin by demonstrating that stochastic VFM forecasting performs better than deterministic regression, emphasizing the importance of explicitly modeling uncertainty in world modeling.
In \cref{subsec:ablation}, we extensively analyze the effect of different diffusion spaces on the sample quality, justifying the importance of optimal autoencoding of VFM features.
Finally, in \cref{subsec:imagen}, we show that VFM auto-encoding is preferable to PCA compression not only in forecasting, but also in image generation. Specifically, we use it to improve the state-of-the-art image generator \cite{ReDi}.


\subsection{Future Forecasting}%
\label{subsec:forecasting}

\method forecasts future VFM features, which we then decode into various modalities such as semantic segmentation, depth, and surface normals.
We assess the quality of these predictions.

We consider DINO~\cite{oquab24dinov2:} as a representative VFM, which allows a direct comparison to DINO-Foresight~\cite{dino-foresight}, a deterministic regression baseline for feature forecasting.
Since their method does not handle variable context lengths, we retrained it with variable-length contexts for a fair comparison, using the official publicly available implementation.
Our training setup follows their protocol, including the same sampling of temporal windows, dataset splits, and preprocessing.

\paragraph{Datasets.}

We run evaluations on \emph{Cityscapes}~\cite{cityscape} and \emph{Kubric MOVi-A}~\cite{kubric}.
\emph{Cityscapes} provides 2{,}975 training and 500 validation \emph{sequences}, each 30 frames at 16~fps with resolution \(1024\times 2048\); following DINO-Foresight, we downsample them to \(224\times 448\) for computational efficiency.
The \(20^\text{th}\) frame comes with dense semantic labels for 19 classes.
\emph{Kubric MOVi-A} contains 9{,}703 training and 250 validation sequences, each of 24 frames at 12~fps and \(256\times 256\) resolution, depicting 3--10 rigid objects moving on a static background with collisions; full per-frame annotations (segmentation, depth, flow, 3D) are available.
For \emph{Cityscapes} we evaluate our predictions on the official validation split.
For \emph{Kubric} we additionally construct an unseen test set of 128 scenes and generate 64 distinct futures per scene by varying initial object velocities while holding initial poses fixed.
The two datasets are complementary: \emph{Cityscapes} offers diverse real-world dynamics but a single annotated future per clip, whereas \emph{Kubric} enables controlled generation of multiple plausible futures under uncertainty.

\paragraph{Benchmark.}

We evaluate \emph{future forecasting} using three modalities: semantic segmentation, depth, and surface normals.
For each forecasting method and modality, we train {DPT}~\cite{ranftl21vision} decoding heads that map predicted VFM features to targets, following the DINO-Foresight~\cite{dino-foresight} protocol.
On \emph{Cityscapes}, we use the official probing heads released by DINO-Foresight for their model; for ours, we train new heads under the same protocol and codebase for a fair comparison.
On \emph{Kubric}, we train new probing heads for {all} methods within the shared implementation framework.

\paragraph{Metrics.}

Following DINO-Foresight, we report semantic segmentation with mIoU over all classes (mIoU-All) and over movable objects only (MO-mIoU\@; \eg, person, rider, car, truck, bus, train, motorcycle, bicycle); depth with AbsRel and \(\delta_1\); and surface normals with mean angular error \(m\) and the percentage of pixels with error \(<\!11.25^\circ\).
On \emph{Kubric}, because the number of instances varies per scene, we evaluate foreground/background segmentation only.
Metric definitions are provided in the Appendix.

To compare fairly with deterministic regression, we use two evaluation protocols for our stochastic model:
(i) \textbf{Mean-of-\(k\)}: average \(k\) sampled futures \emph{in feature space} to obtain a single prediction (an estimate of the conditional mean given the context), then decode once. This matches the \(\ell_2\) regression target.
(ii) \textbf{Best-of-\(k\)}: compute the metric for each of the \(k\) samples and report the best score. On \emph{Cityscapes} there is one ground truth per clip; on \emph{Kubric}, single-frame rollouts have 64 ground-truth futures per scene, whereas rollouts from \(\geq\!2\) frames are effectively deterministic (horizontal velocities are observable), so a single ground truth is used.
We use \(k{=}32\) on \emph{Cityscapes} and \(k{=}64\) on \emph{Kubric} to balance computational efficiency and evaluation accuracy.


\paragraph{Results.}

As shown in \Cref{tab:forecasting_narrow_compressed}, our generative method substantially outperforms the deterministic baseline. The performance gap is largest when uncertainty is highest (\ie, at shorter context lengths), highlighting our key contribution: \textit{explicit uncertainty modeling}. While all methods improve with longer context lengths, our approach consistently achieves the best performance across all evaluation settings, suggesting it can dynamically adapt to the amount of information provided. However, our best-of-\(k\) sometimes underperforms the mean-of-\(k\) estimate, especially on Cityscapes. This reflects the limitation of evaluating against a single ground truth with a limited number of samples (\(k\in\{32,64\}\)). Supporting this interpretation, Kubric, with 64 available ground truths, reveals a much larger gap between best and average performance, particularly at sparse context lengths.


\begin{table}[ht!]
\centering
\caption{\textbf{Dense Future Forecasting Accuracy.} Our \textit{generative} \method considerably outperforms \textit{deterministic} DINO-Foresight on both CityScapes and Kubric, highlighting benefits of explicit modeling of uncertainty in future. }
\label{tab:forecasting_narrow_compressed}
\footnotesize 
\setlength{\tabcolsep}{2pt} 
\sisetup{
  round-mode=places,
  round-precision=2 
}
\begin{tabular}{
  l 
  S[table-format=2.2] 
  S[table-format=2.2] 
  S[table-format=2.2] 
  S[table-format=1.2] 
  S[table-format=2.2] 
  S[table-format=1.2] 
}
\toprule
\multirow{2}{*}{\textbf{Model}} & \multicolumn{2}{c}{\textbf{Segmentation}} & \multicolumn{2}{c}{\textbf{Depth}} & \multicolumn{2}{c}{\textbf{Normals}} \\
\cmidrule(lr){2-3} \cmidrule(lr){4-5} \cmidrule(lr){6-7}
 & {All ($\uparrow$)} & {Mov. ($\uparrow$)} & {d1 ($\uparrow$)} & {AbsRel ($\downarrow$)} & {a3 ($\uparrow$)} & {MeanAE ($\downarrow$)} \\
\midrule

\multicolumn{7}{l}{\textit{Initial Context Length $|\mathcal{C}|$ = 1}} \\
\midrule
\multicolumn{7}{l}{\textbf{CityScapes} (\textit{roll out future 9 frames})} \\
\quad DINO-Foresight & 31.6723200 & 22.0003990 & 70.7711720 & 0.2347890 & 84.8175590 & 5.3349400 \\
\quad \method (Mean) & {\textbf{34.05}} & {\textbf{30.94}} & 75.6294700 & 0.1970750 & 88.1053930 & 4.6296790 \\
\quad \method (Best) & 31.7376110 & 28.5584100 & {\textbf{78.47}} & {\textbf{0.18}} & {\textbf{89.32}} & {\textbf{4.44}} \\
\midrule
\multicolumn{7}{l}{\textbf{Kubric} (\textit{roll out future 11 frames})} \\
\quad DINO-Foresight & 46.728558 & 5.106874 & 64.370214 & 0.24366 & 90.617921 & 2.937612 \\
\quad \method (Mean) & 48.289867 & 6.611382 & 68.326973 & 0.205459 & 93.29122 & 2.144349 \\
\quad \method (Best) & {\textbf{70.55}} & {\textbf{47.77}} & {\textbf{88.80}} & {\textbf{0.08}} & {\textbf{93.45}} & {\textbf{2.07}} \\
\midrule

\multicolumn{7}{l}{\textit{Initial Context Length $|\mathcal{C}|$ = 2}} \\
\midrule
\multicolumn{7}{l}{\textbf{CityScapes} (\textit{roll out future 8 frames})} \\
\quad DINO-Foresight & 39.3517920 & 32.2585910 & 74.3362090 & 0.2074690 & 87.1377460 & 4.8624950 \\
\quad \method (Mean) & {\textbf{41.69}} & {\textbf{38.92}} & 77.8557770 & 0.1819600 & 90.8074190 & 4.1243270 \\
\quad \method (Best) & 39.5664780 & 36.7041590 & {\textbf{80.24}} & {\textbf{0.17}} & {\textbf{91.32}} & {\textbf{4.04}} \\
\midrule
\multicolumn{7}{l}{\textbf{Kubric} (\textit{roll out future 10 frames})} \\
\quad DINO-Foresight & 51.145584 & 14.394726 & 62.600888 & 0.239131 & 89.196849 & 3.361023 \\
\quad \method (Mean) & 55.893142 & 21.676942 & 68.868614 & 0.224827 & \textbf{92.31} & 2.386842 \\
\quad \method (Best) & {\textbf{64.84}} & {\textbf{37.97}} & {\textbf{78.06}} & {\textbf{0.16}} & 91.659134 & {\textbf{2.59}} \\
\midrule

\multicolumn{7}{l}{\textit{Initial Context Length $|\mathcal{C}|$ = 3}} \\
\midrule
\multicolumn{7}{l}{\textbf{CityScapes} (\textit{roll out future 7 frames})} \\
\quad DINO-Foresight & 41.8855950 & 35.6074060 & 75.9108380 & 0.1914300 & 88.4167280 & 4.6049470 \\
\quad \method (Mean) & {\textbf{44.31}} & {\textbf{41.64}} & 79.6406490 & 0.1670230 & 91.6613220 & 3.9466050 \\
\quad \method (Best) & 42.4148920 & 39.6591710 & {\textbf{81.53}} & {\textbf{0.16}} & {\textbf{92.26}} & {\textbf{3.85}} \\
\midrule
\multicolumn{7}{l}{\textbf{Kubric} (\textit{roll out future 9 frames})} \\
\quad DINO-Foresight & 54.278188 & 19.954559 & 66.758659 & 0.223816 & 89.425004 & 3.261334 \\
\quad \method (Mean) & 59.466175 & 27.78518 & 71.934597 & 0.20883 & {\textbf{92.67}} & 2.261647 \\
\quad \method (Best) & {\textbf{68.25}} & {\textbf{43.74}} & {\textbf{80.09}} & {\textbf{0.14}} & 92.313614 & {\textbf{2.40}} \\
\midrule

\multicolumn{7}{l}{\textit{Initial Context Length $|\mathcal{C}|$ = 4}} \\
\midrule
\multicolumn{7}{l}{\textbf{CityScapes} (\textit{roll out future 6 frames})} \\
\quad DINO-Foresight & 44.7466030 & 38.9247470 & 77.6553680 & 0.1768210 & 89.8656370 & 4.3098360 \\
\quad \method (Mean) & {\textbf{46.25}} & {\textbf{43.66}} & 80.5694300 & 0.1584540 & 92.4275580 & 3.7920350 \\
\quad \method (Best) & 44.4885630 & 41.8167650 & {\textbf{82.41}} & {\textbf{0.15}} & {\textbf{92.95}} & {\textbf{3.71}} \\
\midrule
\multicolumn{7}{l}{\textbf{Kubric} (\textit{roll out future 8 frames})} \\
\quad DINO-Foresight & 57.618561 & 25.778604 & 69.308382 & 0.215344 & 89.823544 & 3.113956 \\
\quad \method (Mean) & 61.741413 & 31.857861 & 71.876375 & 0.205571 & {\textbf{92.66}} & 2.247179 \\
\quad \method (Best) & {\textbf{69.86}} & {\textbf{46.56}} & {\textbf{80.50}} & {\textbf{0.14}} & 92.621902 & {\textbf{2.31}} \\
\bottomrule
\end{tabular}
\end{table}
\begin{figure}
    \centering
    \includegraphics[width=0.5\textwidth]{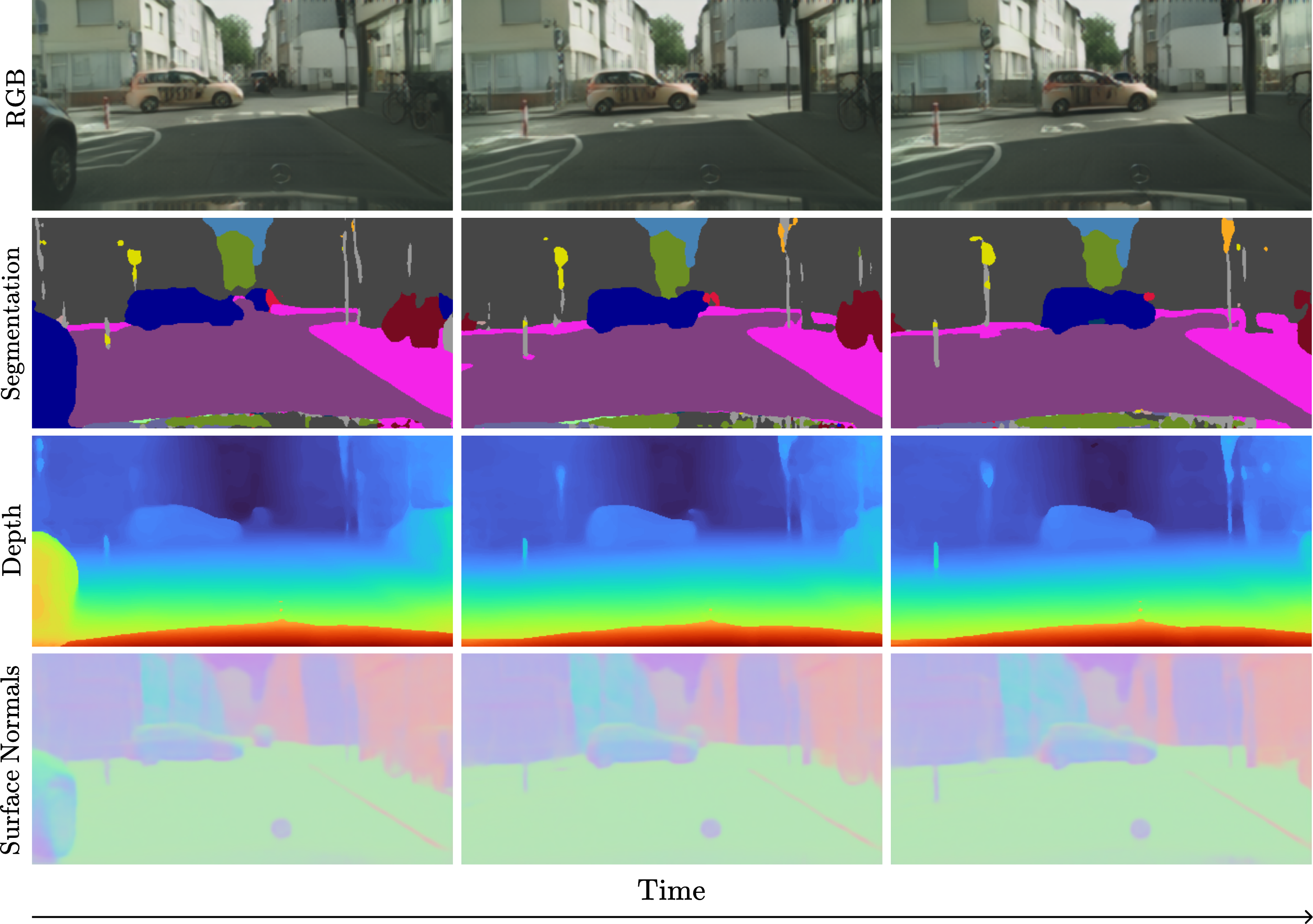}
    \caption{\textbf{One feature set, many modalities:} Our diverse generated futures (\Cref{fig:regressionblurry}) can be translated to 
    diverse modalities, from pixels (RGB) to semantics and geometry (depth, normals).}
    \label{fig:modalities}
\end{figure}

\subsection{How (Not) to Diffuse DINO Features?}%
\label{subsec:ablation}
\begin{figure}
    \includegraphics[width=0.9\linewidth]{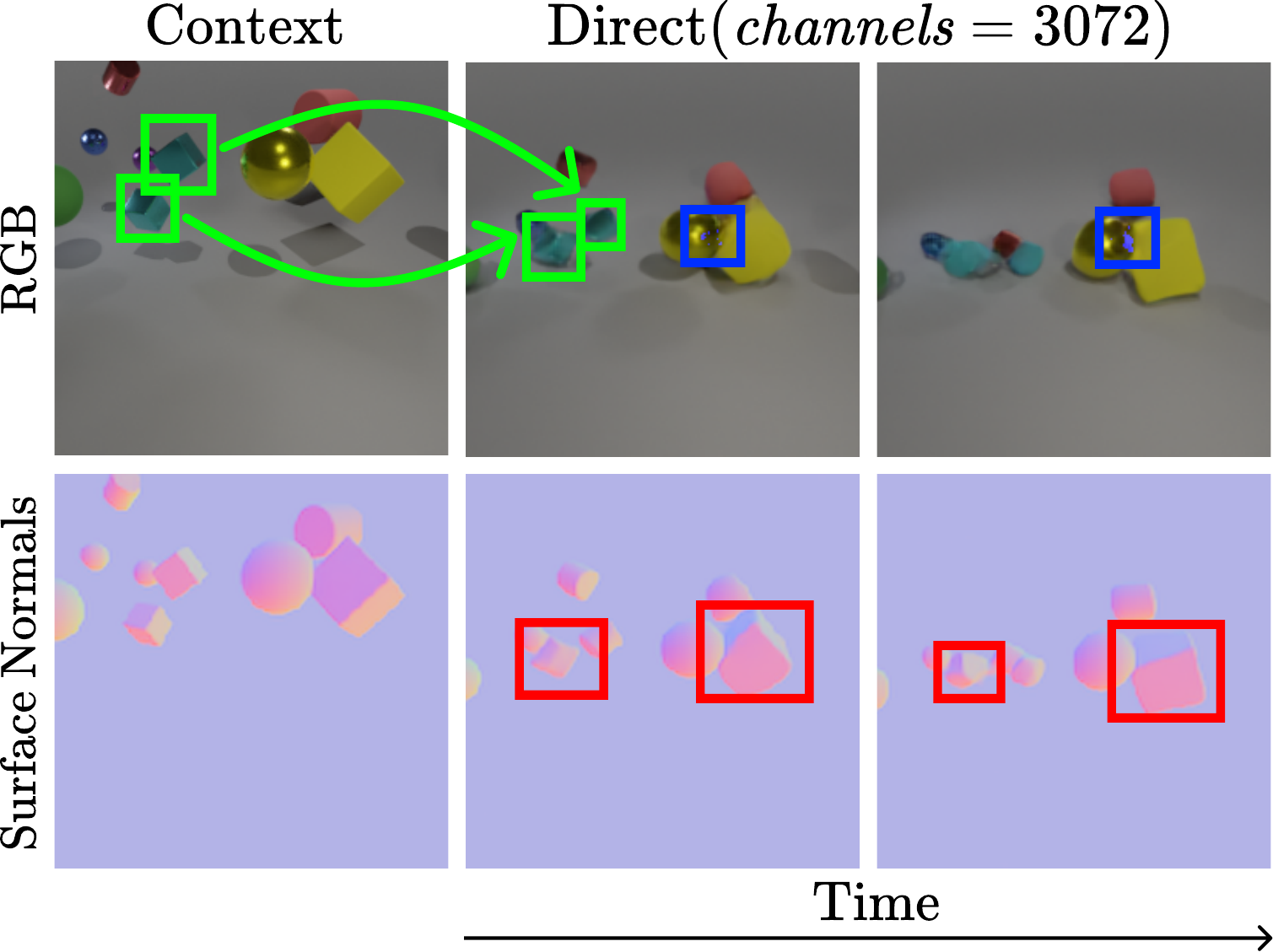}
    \caption{\textbf{Decoded futures} generated by directly diffusing DINO features. Direct diffusion of DINO features leads to
unrealistic motion with several failure modes: \geomdist, \shapeinc, or \rgbart. }
    \label{fig:direct_diffusion}
\end{figure}
\begin{figure*}
    \centering
    \includegraphics[width=\linewidth]{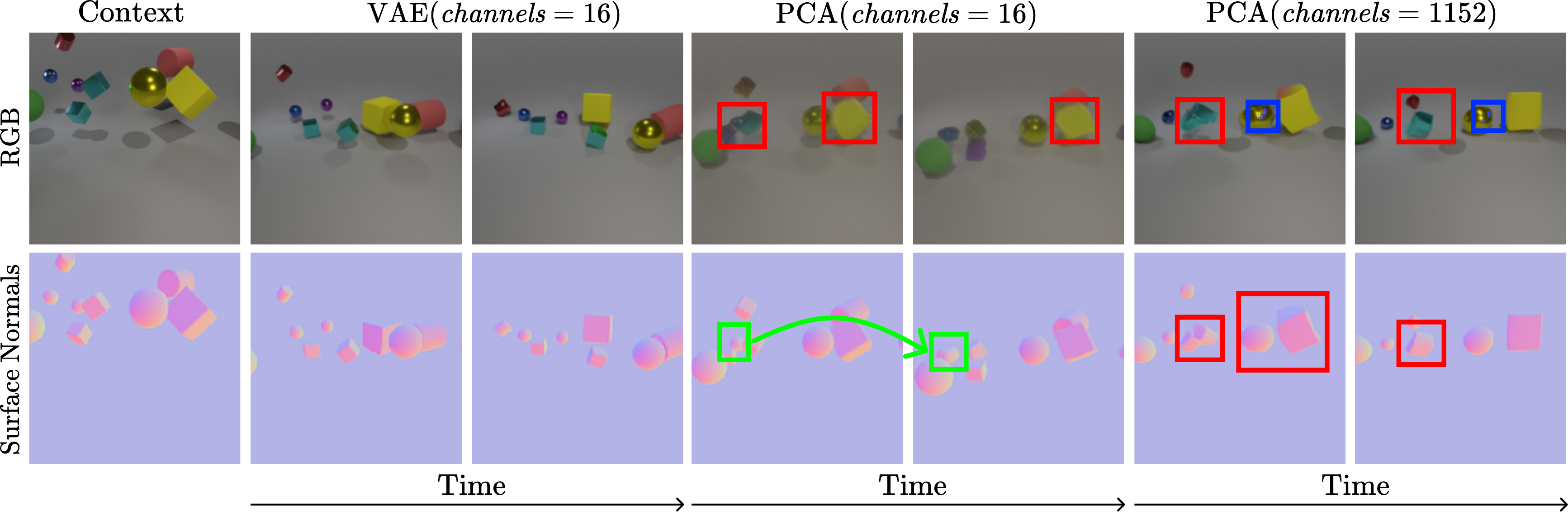}
\caption{\textbf{Qualitative comparison} of future prediction decoded from 
different latent spaces. VAE latent diffusion yields superior prediction quality. 
Alternative methods suffer from either \geomdist, \shapeinc, or \rgbart.}

    \label{fig:vae_vs_pca}
\end{figure*}


Once we frame future forecasting as conditional generation, a straightforward approach is to attempt to diffuse DINO features directly. However, as shown in \cref{fig:direct_diffusion}, this produces unsatisfactory results even on the simple Kubric dataset. Specifically, direct diffusion of DINO features leads to unrealistic motion with several failure modes: distorted object geometry, objects merging together, and non-rigid deformations. When the diffused features are translated to RGB, they also exhibit various artifacts, indicating residual noise resulting from imperfect denoising.

\paragraph{Feature Compression Is Crucial for Generation Quality.} 
First, we hypothesize that the aforementioned artifacts are \textit{primarily} a consequence of the curse of dimensionality - \textit{diffusing thousands of channels, while requiring temporal consistency is too difficult}. Actually, modern image/video diffusion models~\cite{chen2025deepcompressionautoencoderefficient,peng2025opensora20trainingcommerciallevel,wan2025wanopenadvancedlargescale,ai2025magi1autoregressivevideogeneration,hacohen2024ltxvideorealtimevideolatent} diffuse \textit{heavily compressed} RGB latents, by up to \(192\times\). To test this hypothesis, we compress features (by \(192\times\)) with two methods, namely PCA and VAE, while keeping the latent dimension fixed. ~\Cref{fig:vae_vs_pca} shows that diffusing 16 channels significantly outperforms diffusing DINO features directly (\Cref{fig:direct_diffusion}) or PCA compressed with a higher rank (1152 instead of 16). In particular, most of the shape inconsistency artifacts disappear, as further evidenced by sharply decoded geometry (surface normals, depth). This also reflects in significantly improved downstream forecasting accuracy on both datasets. Relevant quantitative evidence is in the supplement. Such a performance gain strongly supports the hypothesis that low-dimensional diffusion is easier.

\begin{table}[ht!]
    \centering
    \footnotesize 
    \caption{\textbf{Dense decoding performance of  decoding heads}, evaluated on ground truth annotated futures. Under equal latent capacity, PCA exhibits severe degradation, while the VAE maintains high reconstruction quality, significantly higher than PCA.}
    \label{tab:heads}
    \setlength{\tabcolsep}{2pt} 
    \sisetup{
      round-mode=places,
      round-precision=2 
    }

    \begin{tabular}{
      l                    
      S[table-format=2.2]    
      S[table-format=2.2]    
      S[table-format=2.2]    
      S[table-format=1.2]    
      S[table-format=2.2]    
      S[table-format=1.2]    
    }
    \toprule
    \multirow{2}{*}{\textbf{Model}} & \multicolumn{2}{c}{\textbf{Segmentation}} & \multicolumn{2}{c}{\textbf{Depth}} & \multicolumn{2}{c}{\textbf{Normals}} \\
    \cmidrule(lr){2-3} \cmidrule(lr){4-5} \cmidrule(lr){6-7}
     & {All ($\uparrow$)} & {Mov. ($\uparrow$)} & {d1 ($\uparrow$)} & {AbsRel ($\downarrow$)} & {a3 ($\uparrow$)} & {MeanAE ($\downarrow$)} \\
    \midrule
    \multicolumn{7}{l}{\textbf{Cityscapes}} \\
    \quad Direct DINO & 68.42170715332031 & 66.80571746826172 & 87.17694282531738 & 0.11312608420848846 & 96.925926 & 2.938775 \\
    \quad PCA (1152) & 68.09736633300781 & 67.27743530273438 & 85.73092818260193 & 0.11500895023345947 & 96.9775390625 & 2.9147424697875977 \\
    \quad PCA (16) & 54.65583801269531 & 49.64780807495117 & 81.5988302230835 & 0.15756788849830627 & 94.76048278808594 & 3.4985973834991455 \\
    \quad VAE(16) & 65.64169311523438 & 64.31041717529297 & 84.75431203842163&  0.12315826117992401& 96.94222259521484 & 2.917991876602173\\
    \midrule
    \multicolumn{7}{l}{\textbf{Kubric}} \\
    \quad Direct DINO & 99.53663635253906 & 99.160400390625 & 77.62729525566101 &  0.15639297664165497 &  99.61978149414062 & 0.28796741366386414 \\
    \quad PCA (1152) & 99.49028778076172 & 99.0764389038086 &  77.51978039741516 & 0.1531732976436615 & 99.61569213867188 & 0.28686249256134033 \\
    \quad PCA (16) & 97.4259033203125 & 95.34408569335938 & 77.45527029037476& 0.15866990387439728 & 99.06644439697266& 0.4457050561904907 \\
    \quad VAE(16)  & 99.16780853271484 & 98.49254608154297 &  78.01956534385681 &  0.15711060166358948& 99.55836486816406 &  0.302557110786438 \\
    \bottomrule
    \end{tabular}

\end{table}
\paragraph{PCA Compression Is Suboptimal.} Although low-rank PCA simplifies latent diffusion, it incurs noticeable information loss. Specifically, \Cref{fig:vae_vs_pca} shows that while high-level properties such as semantics and geometry (normals) are well-preserved, pixel-level details are lost—which is crucial for high-fidelity RGB synthesis (\cref{fig:imagen}).

To quantify this information loss, we train modality-specific decoders on autoencoded features from each  method and evaluate their accuracy on ground truth data. \Cref{tab:heads} presents the results, indicating that PCA performs significantly worse than VAE on downstream dense prediction tasks. 

We further demonstrate the advantages of VAE in \Cref{subsec:imagen}, where we show that our VAE improves ReDi~\cite{ReDi}, the state-of-the-art method for joint DINO PCA and RGB image generation.
 
It is worth noting that high-rank PCA preserves fine-grained details but reintroduces geometry artifacts similar to those observed with direct diffusion (\cref{fig:direct_diffusion}). In other words, naively increasing latent capacity preserves information at the expense of generation quality~\cite{yao2025vavae}, justifying the need for a more sophisticated autoencoder.

\paragraph{Autoencoding DINO Requires Careful Spectral Analysis.}

Recently, the works of~\cite{skorokhodov2025improvingdiffusabilityautoencoders, kouzelis2025eqvaeequivarianceregularizedlatent} show that modern RGB autoencoders over-represent high-frequency information in their latent spaces.
This creates a mismatch with the \textit{coarse-to-fine nature}~\cite{dieleman2024spectral, wang2025ddtdecoupleddiffusiontransformer, falck2025fourierspaceperspectivediffusion} of denoising diffusion, harming sample quality. Equalizing the frequency spectra of the latent and RGB modalities substantially improves generation quality~\cite{kouzelis2025eqvaeequivarianceregularizedlatent,skorokhodov2025improvingdiffusabilityautoencoders}.
Motivated by these results, we ask: ``Do these findings from the RGB domain transfer to DINO features?''.

\begin{figure}[h!]
    \centering
    \input{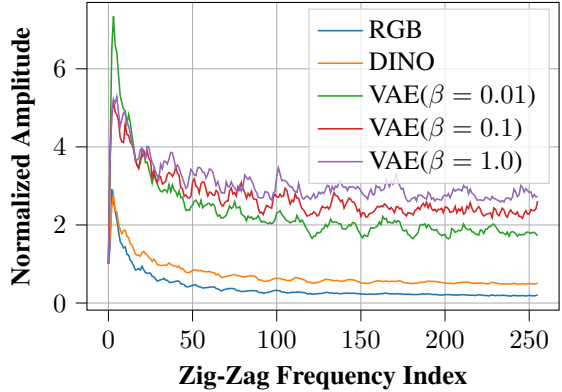}  
    \vspace{-1em}
    \caption{\textbf{Frequency profiles} on Kubric. Uncompressed DINO features exhibit spectral characteristics similar to RGB inputs. As Gaussian regularization on the compressed features increases, the spectrum shifts toward higher frequencies, reflecting noise injected into the latent space.}
    \label{fig:spectrum}
\end{figure}

To this end, we first perform spectral decomposition of DINO features or their latents, quantifying the power of each DCT basis function sorted by its \textit{zig-zag} frequency. \Cref{fig:spectrum} shows that DINO features exhibit spectral power laws similar to RGB. 
Prior works ~\cite{karras2024analyzingimprovingtrainingdynamics,chen2023importancenoiseschedulingdiffusion} have noted the crucial importance of \textit{signal-to-noise} ratio (SNR), heavily affected by latent scale. However, naively increasing KL regularization to constrain the latent scale shifts the spectrum away from RGB, reflecting the higher amount of noise in latents. This causes mismatch with the \textit{coarse-to-fine nature}~\cite{dieleman2024spectral, wang2025ddtdecoupleddiffusiontransformer, falck2025fourierspaceperspectivediffusion} of denoising diffusion, requiring careful selection of VAE hyperparameters that balances reconstruction ability, latent scale, and spectral properties. Concretely, we use $\beta=0.01$ in all experiments.

\subsection{Guiding Image Generation with \method\ VAE}%
\label{subsec:imagen}

We show the benefits of our VFM-VAE by utilizing it in the ReDi~\cite{ReDi} image generator model.
Recall that ReDi uses PCA to compress DINO features as an additional target for generation, along with the RGB modality.
They show, remarkably, that predicting RGB+DINO jointly improves the image generation quality, related to the findings of~\cite{RCG2023}.
However, while they use a standard latent space for RGB, they employ PCA to reduce the dimensionality of DINO\@.
We instead train a VAE of the same dimensionality on \emph{ImageNet}~\cite{deng09imagenet:} and use the resulting VAE-compressed features as a drop-in replacement.
For fairness, we train both variants from scratch with the SiT backbone~\cite{ma2024sitexploringflowdiffusionbased} at two scales (SiT-B, SiT-XL), matching data, optimizer, and budget (400k updates), evaluating with the standard ADM~\cite{karras2024analyzingimprovingtrainingdynamics} protocol. We evaluate quality of compressed features by applying \textit{representation guidance} ~\cite{ReDi} during sampling.

\paragraph{Results.}

\Cref{tab:imagen} reports consistent improvements for our VAE-guided variant over the PCA-guided baseline at 400k steps, across SiT-B and SiT-XL and for multiple values of the VAE KL weight $\beta$.
Qualitative comparisons on SiT-XL (\cref{fig:imagen}) show sharper textures and better semantic faithfulness when guiding with VAE-compressed VFM features.
Moreover, on SiT-B our variant converges faster (\cref{fig:convergence}), indicating that VAE-DINO provides a better guidance signal than PCA-DINO\@.

\begin{table}[htbp]
\centering
\small 
\setlength{\tabcolsep}{3pt} 
\renewcommand{\arraystretch}{0.95} 

\sisetup{round-mode=places, round-precision=2, detect-weight=true}

\caption{\textbf{Image quality on conditional ImageNet 256×256.} 
Results are reported with optimal representation guidance scale $w_r$, after 400K training iterations. 
Diffusing our VAE latents instead of PCA projections results in higher-quality samples that better match the diversity of the ground truth distribution.}
\label{tab:imagen}

\begin{tabular}{@{}l
                S[table-format=2.2]
                S[table-format=1.2]
                S[table-format=1.2]
                S[table-format=1.2]@{}}
\toprule
\textbf{Method} & {\textbf{FID} ($\downarrow$)} & {\textbf{sFID} ($\downarrow$)} & {\textbf{Prec.} ($\uparrow$)} & {\textbf{Rec.} ($\uparrow$)} \\
\midrule
\textit{SiT-B} & & & & \\
ReDi (PCA) {\cite{ReDi}} & 18.493 & 6.333 & 0.584 & 0.653 \\
ReDi (VAE, $\beta{=}0.01$) & \bfseries 11.759 & 5.526 & 0.576 & \bfseries 0.728 \\
ReDi (VAE, $\beta{=}0.001$) & 12.634 & \bfseries 5.374 & \bfseries 0.596 & 0.705 \\
\midrule
\textit{SiT-XL} & & & & \\
ReDi (PCA) {\cite{ReDi}} & 5.475 & 4.661 & 0.591 & 0.767 \\
ReDi (VAE, $\beta{=}0.01$) & 5.005 & \bfseries 4.477 & 0.609 & 0.766 \\
ReDi (VAE, $\beta{=}0.001$) & \bfseries 4.978 & 4.545 & \bfseries 0.611 & \bfseries 0.767 \\
\bottomrule
\end{tabular}
\end{table}
\begin{figure}
    \centering
    \includegraphics[width=\linewidth]{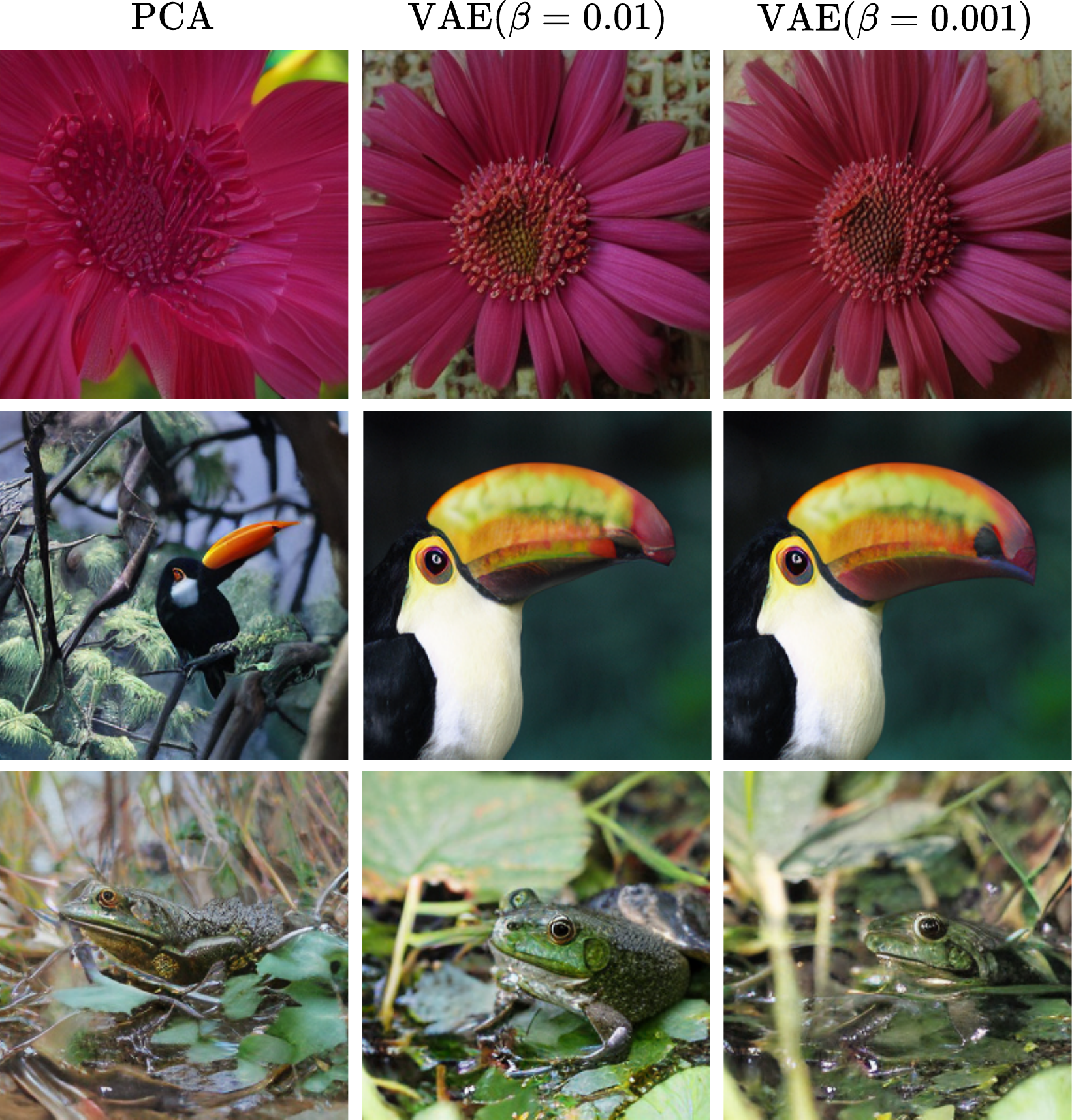}
    \caption{\textbf{Qualitative comparison of image quality} of \textbf{SiT-XL}, with ReDi guidance at 400K training steps. Diffusing VAE latents instead of PCA projections enhances fidelity, realism, and sharpness, resulting in higher quality samples.}
    \label{fig:imagen}
\end{figure}
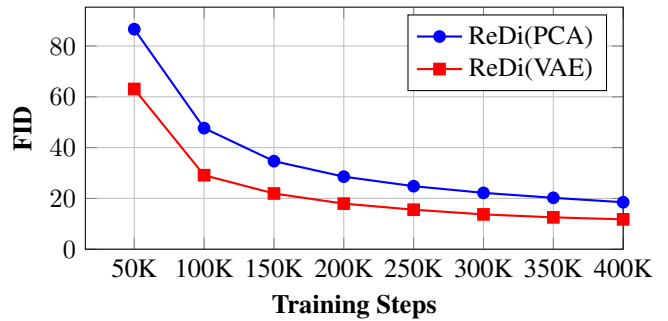
\begin{figure}
    \centering
    \begin{tikzpicture}
        \begin{axis}[
            xlabel={\textbf{Training Steps}},
            ylabel={\textbf{FID}},
            width=0.5\textwidth,
            height=0.275\textwidth,
            grid=major,
            legend pos=north east,
            xmax=400000,
            ymin=0, 
            enlarge y limits={upper, value=0.1}, 
            xtick={50000, 100000, 150000, 200000, 250000, 300000, 350000, 400000},
            xticklabels={50K, 100K, 150K, 200K, 250K, 300K, 350K, 400K},
            scaled x ticks=false, 
        ]
        \addplot[mark=*, color=blue, thick] coordinates {
            (50000, 86.6396999)
            (100000, 47.66971024)
            (150000, 34.6757324)
            (200000, 28.56877841)
            (250000, 24.81495996)
            (300000, 22.18977025)
            (350000, 20.23456498)
            (400000, 18.49305575)
        };
        \addlegendentry{ReDi(PCA)}
        \addplot[mark=square*, color=red, thick] coordinates {
            (50000, 63.08489057)
            (100000, 29.13829599)
            (150000, 21.94340577)
            (200000, 17.94953168)
            (250000, 15.56555068)
            (300000, 13.71012824)
            (350000, 12.54932724)
            (400000, 11.75866361)
        };
        \addlegendentry{ReDi(VAE)}
        \end{axis}
    \end{tikzpicture}
\vspace{-2em}

    \caption{\textbf{FID on conditional ImageNet 256x256.} Replacing PCA ($c=8$) projections with our  VAE latents ($c=8,\beta=0.01$) in ReDi~\cite{ReDi}, applied to SiT-B~\cite{ma2024sitexploringflowdiffusionbased}, yields faster convergence and consistently better generation quality.
    }
    \label{fig:convergence}
\end{figure}

\section{Conclusion}%
\label{sec:conclusion}
We study world modeling with variable-length contexts and show that deterministic regression in VFM feature space averages over uncertain futures, degrading accuracy. We address this by generating autoregressively in a compact latent space of VFM features, yielding uncertainty-aware, sharper predictions at comparable compute. Across multiple modalities, our method outperforms regression baselines, suggesting that stochastic generation of VFM features is a promising foundation for scalable world models.

\paragraph{Acknowledgments.} We thank Isambard-AI and Dawn AIRR supercomputers for supporting this project.

\clearpage

\clearpage
\appendix
\setcounter{page}{1}
\maketitlesupplementary

\label{sec:supplementary}
\begin{figure*}[!h]
    \centering
    \includegraphics[width=\linewidth]{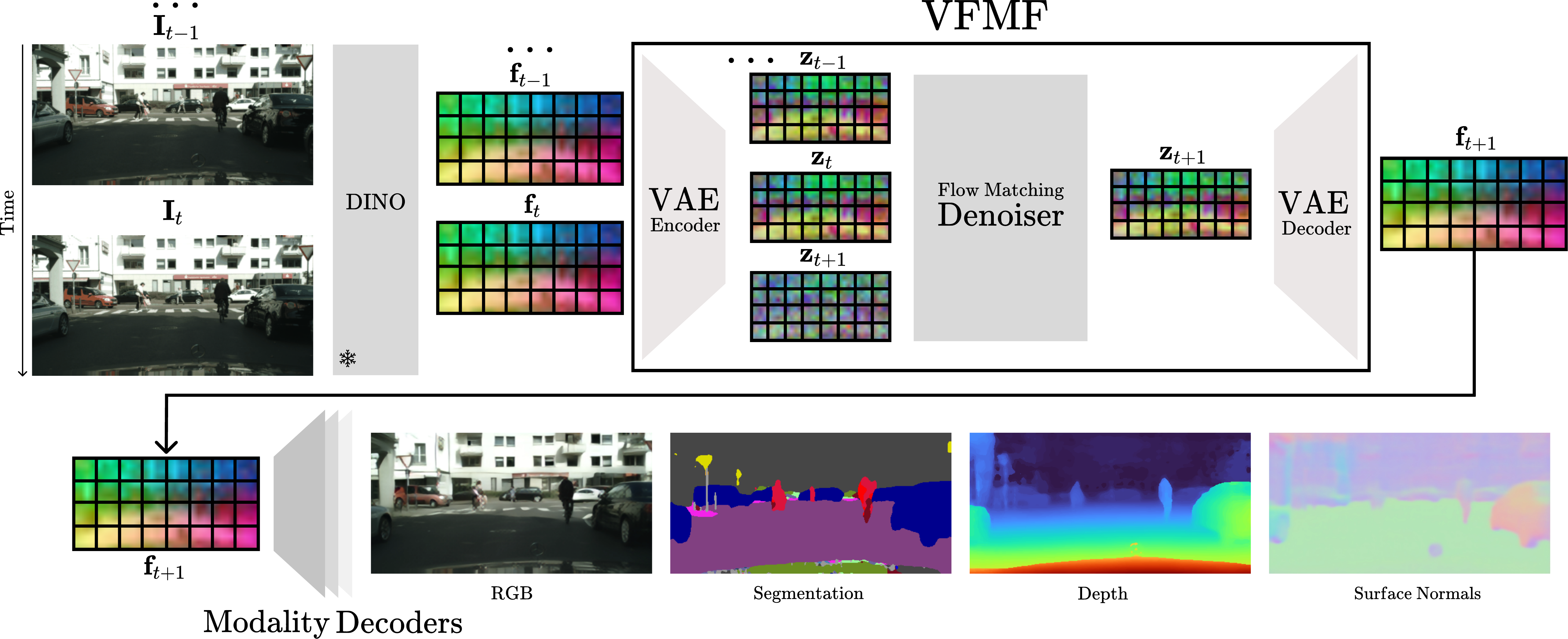}
\caption{\textbf{An overview of our method VFMF.} Given RGB context frames $\mathbf{I}_1, \dots, \mathbf{I}_t$, we extract DINO features $\mathbf{f}_1, \dots, \mathbf{f}_t$ and predict the next state feature $\mathbf{f}_{t + 1}$. Context features are compressed with a VAE along the channel dimension to produce context latents $\mathbf{z}_1, \dots, \mathbf{z}_t$. Those context latents are concatenated with noisy future latents $\mathbf{z}_{t+1}$ and passed to a conditional denoiser that denoises only the future latents $\mathbf{z}_{t+1}$ while leaving the context latents unchanged. This process repeats autoregressively, with a window of fixed length. Specifically, each time a new latent $\mathbf{z}_{t+1}$ is generated, it is appended to the context while the oldest context latent is popped. The denoised future latents are decoded back to DINO feature space by the VAE decoder. Finally, the reconstructed features can be routed to task-specific modality decoders for downstream tasks or interpretation.}
    \label{fig:method}
\end{figure*}

In this supplementary material, we provide: 

\begin{enumerate}
    \item More details about our method and its implementation, architecture, training/sampling hyperparameters in \Cref{sec:impl}.
    \item Performance of different latent spaces on downstream tasks, together with spectral analysis in \Cref{sec:ablations}.
    \item Qualitative examples of feature forecasting (\textbf{offline website} in the \texttt{samples} folder of the supplementary material archive) and image generation samples (ImageNet $256\times256$) in \Cref{sec:qualitative}.
    \item Discussion on limitations and future work in \Cref{sec:limitations}.
\end{enumerate}


\section{Implementation Details}
\label{sec:impl}

An overview of our method is presented in \Cref{fig:method}.

\paragraph{Multi-Scale Feature Space}
It is well known that different layers of foundation models capture information at different scales \cite{bolya2025perceptionencoderbestvisual, carreira25scaling, amir2022deepvitfeaturesdense}. In particular, \cite{bolya2025perceptionencoderbestvisual} demonstrates that the most informative visual features in vision foundation models (VFMs) often reside in intermediate layers. Motivated by this observation, we also work with features from multiple layers of VFM.

Specifically, we use the DINOv2~\cite{oquab24dinov2:} ViT-B(\textit{with registers~\cite{darcet2023vitneedreg}}), extracting and concatenating features from layers \texttt{$3,6,9,12$}, following DINO-Foresight~\cite{dino-foresight}. While the resulting multi-scale feature space increases expressivity, it also substantially raises feature dimensionality. For instance, concatenated ViT-B features are $3072$ dimensional. 
The concatenated features are further compressed by our proposed VAE to mitigate the curse of dimensionality.

\paragraph{Feature VAE} Our VAE employs an encoder-decoder architecture with shared design principles, performing compression exclusively along the channel dimension. We explore both convolutional and transformer-based architectures in three sizes: S\textit{mall}, B\textit{ase}, and L\textit{arge}. The convolutional VAEs build on the \textit{isotropic} ConvNeXt architecture \cite{liu2022convnet2020s}, while the transformer variants are based on the feature transformer from DINO-Foresight \cite{dino-foresight}. Details about these architectures are in~\Cref{tab:cnn_hyperparams_full,tab:vit_hyperparams_full}. For compute constraints, we use convolutional VAEs since they require significantly less GPU memory to train. 

The encoder operates as follows: it takes an input feature map $\vf \in \mathbb{R}^{H\times W \times D}$ and projects it linearly to the model dimension $\vf \in \mathbb{R}^{H\times W \times D_{\operatorname{model}}}$. This representation is then processed through a sequence of isotropic layers (either transformer or convolutional blocks). Finally, a linear projection maps from $\mathbb{R}^{H\times W \times D_{\operatorname{model}}}$ to $\vz \in \mathbb{R}^{H\times W \times 2 \times D_{\operatorname{latent}}}$, producing the mean and log variance of the latent distribution.

The decoder mirrors the encoder architecture. It begins by projecting the latent representation $\vz \in \mathbb{R}^{H\times W \times \times D_{\operatorname{latent}}}$ back to the model dimension $\bm{h} \in \mathbb{R}^{H\times W \times D_{\operatorname{model}}}$ via a linear projection. This representation is then processed through a sequence of isotropic layers (either transformer or convolutional blocks). Finally, a linear projection maps from $\mathbb{R}^{H\times W \times D_{\operatorname{model}}}$ to $\hat{\vf} \in \mathbb{R}^{H\times W \times D}$.

The number of isotropic layers is the same in the encoder and decoder. For a fair comparison with ReDi~\cite{ReDi}, we set our VAE’s latent dimensionality to match ReDi’s PCA rank ($8$ channels). Here, we also used only the final layer features same as ReDi. For a fair comparison with DINO-Foresight, we used a 16-channel VAE and considered PCA baselines with a rank of $16$ or the official DINO-Foresight setting of a rank of $1152$.

\textbf{Training.} We train all models using the \texttt{AdamW} optimizer \cite{kingma2017adammethodstochasticoptimization,loshchilov2019decoupledweightdecayregularization} with a learning rate of $3 \times 10^{-4}$ and linear warmup, gradient clipping at norm 1.0, and mixed precision (\texttt{bfloat16}). For Cityscapes and Kubric, we use an effective batch size of 256 across $4$ or $8$ L40s 40GB GPUs, training for 200 epochs on Cityscapes and 2000 epochs on Kubric (approximately 2 days each). For ImageNet, we train for 100 epochs. We use an effective batch size of 2048 distributed across 4 NVIDIA GH200 Grace Hopper Superchips. This takes approximately 1 day.

\paragraph{Feature Denoiser}
Our denoising network architecture builds upon the masked feature transformer from DINO-Foresight. The network uses 12 transformer layers with a hidden size of 1152. Each input sequence contains up to 4 context frames ($|\mathcal{C}| = 4$) and 1 noisy prediction frame, where the network denoises only the prediction frame.

We extend the original DINO-Foresight architecture with two key components for denoising: (1) timestep encoding and (2) timestep injection.

\textbf{Timestep Encoding.} We introduce a flow matching time input $t \in [0,1]$, which is processed through standard sinusoidal positional encoding with 256 frequencies. The encoded timestep is then projected to the transformer dimension via a 2-layer MLP.

\textbf{Timestep Injection.} Following DiT~\cite{peebles23scalable} and SiT~\cite{ma2024sitexploringflowdiffusionbased}, we condition the network on the timestep embedding using zero-adaptive normalization (adaLN-Zero). Specifically, the timestep embedding passes through another 2-layer MLP that regresses 9 adaptive normalization parameters: shift, scale, and gate parameters for spatial attention, temporal attention, and the MLP. These parameters are shared across all transformer blocks, as in~\cite{chen2023pixartalpha,gao2024lumin-t2x,lan2024ga}.

Since DINO-Foresight employs spatial-temporal attention, we replicate the regressed normalization parameters along both spatial and temporal dimensions. The adaptive normalization is then applied after each spatial attention, temporal attention, and MLP operation within every transformer block. Finally, as in DiT and SiT, we add a final adaptive normalization layer after all transformer blocks, which regresses its own set of parameters.

\textbf{Training.} We use \texttt{AdamW} optimizer \cite{kingma2017adammethodstochasticoptimization,loshchilov2019decoupledweightdecayregularization} with a learning rate of $6.4 \times 10^{-4}$ with cosine annealing. Training is performed on $8$ L40s 40GB GPUs, giving an effective batch size of 64. On Cityscapes, we train for 3200 epochs. On Kubric, we train for 2400 epochs. When training with variable context length, especially on Kubric, we encountered training instability with both DINO-Foresight and our modified architecture. The problem was due to exploding attention logits that we resolved with \texttt{QKNorm}~\cite{henry2020querykeynormalizationtransformers}. 
For fairness, we apply \texttt{QKNorm} to both DINO-Foresight and our method.

\textbf{Sampling}. We use the Euler solver with 10 NFEs for all the experiments.

\paragraph{Probing heads} 
We adhere closely to the DINO-Foresight protocol. While we utilize the official training configuration for CityScapes (base learning rate of $10^{-4}$), we have observed training instability on Kubric. To address this, we decreased the base learning rate on Kubric to $10^{-5}$ for cosine annealing while keeping all other parameters unchanged.

We detail the architectures and training setups used for the various downstream tasks. For semantic segmentation, depth estimation, and surface normal prediction, we employ the DPT head~\cite{ranftl16dense}. We set the feature dimension to 256 and configure 
\texttt{dpt\_out\_channels} = [128, 256, 512, 512].

All models are trained for 100 epochs with an effective batch size of 128 distributed across either 4 or 8 GPUs. Optimization is performed using AdamW with a learning rate of $1.6 \times 10^{-3}$, linear warmup during the first 10 epochs, and a weight decay of $10^{-4}$. We tailor the loss functions and schedulers to each task. \\ \textit{Semantic Segmentation:} We apply a polynomial learning-rate scheduler and optimize with cross-entropy over 19 classes (CityScapes) or 2 classes (Kubric). \\ \textit{Depth Estimation:} We use a cosine annealing scheduler and cross-entropy loss with 256 classes. \\ \textit{Surface Normal Prediction:} We use a polynomial scheduler and a loss combining cosine similarity and $L_2$ distance with weighted averaging over 3 classes.

\begin{table}[h]
\centering
\caption{\textbf{Hyperparameters for ViT (Transformer) VAE Architecture}.}
\label{tab:vit_hyperparams_full}
\begin{tabular}{@{}lcc@{}}
\toprule
\textbf{Hyperparameter} & \textbf{Base} (B) & \textbf{Large} (L) \\
\midrule
dinov2\_variant        & "vitb14\_reg"     & "vitb14\_reg"     \\
intermediate\_layers    & [2, 5, 8, 11]   & [2, 5, 8, 11]   \\
patch\_size             & 14              & 14              \\
input\_dim              & 3072            & 3072            \\
latent\_channels        & 16              & 16              \\
dropout                & 0.1             & 0.1             \\
use\_qk\_norm            & true            & true            \\
abs\_pos\_enc            & true            & true            \\
\midrule
num\_encoder\_layers     & 12              & 24              \\
num\_decoder\_layers     & 12              & 24              \\
heads                  & 12              & 16              \\
hidden\_dim             & 768             & 1024            \\
mlp\_dim                & 3072            & 4096            \\
num\_registers          & 4               & 4               \\
\bottomrule
\end{tabular}
\end{table}

\begin{table}[h]
\centering
\setlength{\tabcolsep}{1pt} 
\caption{\textbf{Hyperparameters for ConvNeXt Isotropic VAE Architecture}.}
\label{tab:cnn_hyperparams_full}
\begin{tabular}{@{}lccc@{}}
\toprule
\textbf{Hyperparameter} & \textbf{Small} (S) & \textbf{Base}(B) & \textbf{Large} (L) \\
\midrule
dinov2\_variant        & "vitb14\_reg" & "vitb14\_reg" & "vitb14\_reg" \\
intermediate\_layers    & [2,5,8,11] & [2,5,8,11] & [2,5,8,11] \\
patch\_size             & 14 & 14 & 14 \\
input\_dim              & 3072 & 3072 & 3072 \\
latent\_channels        & 16 & 16 & 16 \\
drop\_path\_rate        & 0 & 0 & 0 \\
layer\_scale\_init\_value & 0 & 0 & 0 \\
\midrule
depth                  & 18 & 18 & 36 \\
dim                    & 384 & 768 & 1024 \\
\bottomrule
\end{tabular}
\end{table}

\paragraph{RGB Decoder} The RGB-Decoder uses a transformer backbone followed by a DPT-Head~\cite{ranftl16dense} with default parameters. The transformer backbone is identical to the DINO-Foresight ViT encoder (\Cref{tab:vit_hyperparams_full}), using the \textit{base} configuration for Kubric and CityScapes experiments and the \textit{large} configuration for ImageNet. Features are extracted from transformer backbone layers \texttt{[2, 5, 8, 11]} and processed by the DPT head using the VGG-T~\cite{wang25vggt} implementation, resulting in 3 RGB channels. The decoder is trained with an equally-weighted combination of L1 and LPIPS.

\paragraph{Metrics}
We use the same metrics as DINO-Foresight \cite{dino-foresight}, computed using their official code. Please see Section 4.1 in their paper.

\subsection{Image generation}

\textbf{Our method.} We use the official publicly released code from ReDi \cite{ReDi} with a single modification: we replaced PCA with our VAE. The number of VAE channels matches the PCA rank. Additionally, our VAE compresses the same DINO features as PCA. Specifically, for a fair comparison with ReDi, we avoid the multi-scale feature space and use only final layer features.

\textbf{Training.} We use the official publicly released code from ReDi~\cite{ReDi}, removing gradient checkpointing since our hardware supported training without it. All hyperparameters remain identical to those in the original paper. However, we use different hardware: 4 NVIDIA GH200 Grace Hopper Superchips instead of 8 A100 (40GB) GPUs.

\textbf{Sampling.} We follow the same sampling procedure as ReDi~\cite{ReDi}, with official hyperparameters from the paper.

\textbf{Evaluation.} We follow the evaluation approach from ReDi~\cite{ReDi}. Since the paper does not report SiT-B or SiT-XL results with representation guidance at 400K training steps, we trained both models from scratch using the official PCA checkpoint and training protocol, incorporating the modifications described above. We performed a hyperparameter sweep over representation guidance strength $w_r \in \{1.1, 1.2, 1.5\}$ (values from the original paper) for all methods and reported the best results.
\section{Design Choices}
\label{sec:ablations}

\paragraph{How to diffuse DINO features?}

\Cref{tab:cityscapes_latent_space_ablation,tab:kubric_latent_space_ablation} show that diffusing VAE latents outperforms the alternatives across a wide range of downstream tasks.

\begin{table*}[h!]
\centering
\caption{\textbf{Dense Forecasting Accuracy} with different diffusion spaces on CityScapes. The VAE latent diffusion consistently delivers the best overall performance for dense forecasting.}
\label{tab:cityscapes_latent_space_ablation}

\renewcommand{\arraystretch}{0.85}

\resizebox{!}{0.475\textheight}{%
    \sisetup{round-mode=places, round-precision=3}
    \begin{tabular}{
    l 
    c 
    c 
    c 
    c 
    c 
    c 
    }
    \toprule
    \multirow{2}{*}{\textbf{Model}} & \multicolumn{2}{c}{\textbf{Segmentation (mIoU)}} & \multicolumn{2}{c}{\textbf{Depth}} & \multicolumn{2}{c}{\textbf{Normals}} \\
    \cmidrule(lr){2-3} \cmidrule(lr){4-5} \cmidrule(lr){6-7}
     & {All ($\uparrow$)} & {Mov. ($\uparrow$)} & {d1 ($\uparrow$)} & {AbsRel ($\downarrow$)} & {a3 ($\uparrow$)} & {MeanAE ($\downarrow$)} \\
    \midrule
    
    \multicolumn{7}{l}{\textit{Initial Context Length $|\mathcal{C}|$ = 1, roll out 9 frames}} \\
    \midrule
    DINO-Foresight & 31.672 & 22.000 & 70.771 & 0.235 & 84.818 & 5.335 \\
    \addlinespace
    \multicolumn{7}{l}{\textbf{VAE (L, 16 channels)}} \\
    \quad \method (Mean) & 34.049 & 30.937 & 75.629 & 0.197 & 88.105 & 4.630 \\
    \quad \method (Best) & 31.738 & 28.558 & \textbf{78.471} & \textbf{0.181} & \textbf{89.318} & \textbf{4.435} \\
    \addlinespace
    \multicolumn{7}{l}{\textbf{PCA (16 channels)}} \\
    \quad \method (Mean) & 30.601 & 27.313 & 72.398 & 0.207 & 87.810 & 4.777 \\
    \quad \method (Best) & 30.163 & 26.916 & 77.607 & 0.188 & 89.035 & 4.567 \\
    \addlinespace
    \multicolumn{7}{l}{\textbf{PCA (1152 channels)}} \\
    \quad \method (Mean) & 34.870 & 31.851 & 74.327 & 0.232 & 86.007 & 5.072 \\
    \quad \method (Best) & 33.789 & 30.729 & 76.425 & 0.214 & 86.938 & 4.912 \\
    \addlinespace
    \multicolumn{7}{l}{\textbf{Direct  (3072 channels)}} \\
    \quad \method (Mean) & \textbf{35.044} & \textbf{32.040} & 75.257 & 0.226 & 86.425 & 4.992 \\
    \quad \method (Best) & 33.722 & 30.666 & 77.112 & 0.212 & 87.536 & 4.852 \\
    \midrule
    \multicolumn{7}{l}{\textit{Initial Context Length $|\mathcal{C}|$ = 2, roll out 8 frames}} \\
    \midrule
    DINO-Foresight & 39.352 & 32.259 & 74.336 & 0.207 & 87.138 & 4.862 \\
    \addlinespace
    \multicolumn{7}{l}{\textbf{VAE (L, 16 channels)}} \\
    \quad \method (Mean) & \textbf{41.687} & \textbf{38.920} & 77.856 & 0.182 & 90.807 & 4.124 \\
    \quad \method (Best) & 39.566 & 36.704 & \textbf{80.242} & \textbf{0.167} & \textbf{91.318} & \textbf{4.039} \\
    \addlinespace
    \multicolumn{7}{l}{\textbf{PCA (16 channels)}} \\
    \quad \method (Mean) & 36.337 & 33.284 & 75.242 & 0.193 & 89.645 & 4.442 \\
    \quad \method (Best) & 35.720 & 32.665 & 78.953 & 0.178 & 90.513 & 4.293 \\
    \addlinespace
    \multicolumn{7}{l}{\textbf{PCA (1152 channels)}} \\
    \quad \method (Mean) & 36.942 & 34.002 & 75.540 & 0.218 & 86.713 & 4.936 \\
    \quad \method (Best) & 36.240 & 33.275 & 77.480 & 0.203 & 87.518 & 4.789 \\
    \addlinespace
    \multicolumn{7}{l}{\textbf{Direct  (3072 channels)}} \\
    \quad \method (Mean) & 38.960 & 36.104 & 77.328 & 0.201 & 87.925 & 4.709 \\
    \quad \method (Best) & 37.465 & 34.545 & 79.066 & 0.191 & 88.745 & 4.624 \\
    \midrule
    \multicolumn{7}{l}{\textit{Initial Context Length $|\mathcal{C}|$  = 3, roll out 7 frames}} \\
    \midrule
    DINO-Foresight & 41.886 & 35.607 & 75.911 & 0.191 & 88.417 & 4.605 \\
    \addlinespace
    \multicolumn{7}{l}{\textbf{VAE (L, 16 channels)}} \\
    \quad \method (Mean) & \textbf{44.313} & \textbf{41.640} & 79.641 & 0.167 & 91.661 & 3.947 \\
    \quad \method (Best) & 42.415 & 39.659 & \textbf{81.531} & \textbf{0.157} & \textbf{92.259} & \textbf{3.848} \\
    \addlinespace
    \multicolumn{7}{l}{\textbf{PCA (16 channels)}} \\
    \quad \method (Mean) & 38.585 & 35.623 & 76.427 & 0.187 & 90.416 & 4.300 \\
    \quad \method (Best) & 37.728 & 34.746 & 79.805 & 0.172 & 91.207 & 4.159 \\
    \addlinespace
    \multicolumn{7}{l}{\textbf{PCA (1152 channels)}} \\
    \quad \method (Mean) & 38.955 & 36.086 & 76.902 & 0.207 & 87.797 & 4.728 \\
    \quad \method (Best) & 38.132 & 35.230 & 78.800 & 0.193 & 88.537 & 4.597 \\
    \addlinespace
    \multicolumn{7}{l}{\textbf{Direct  (3072 channels)}} \\
    \quad \method (Mean) & 41.579 & 38.823 & 78.720 & 0.189 & 89.175 & 4.465 \\
    \quad \method (Best) & 40.176 & 37.360 & 80.595 & 0.178 & 89.920 & 4.388 \\
    \midrule
    \multicolumn{7}{l}{\textit{Initial Context Length $|\mathcal{C}|$  = 4, roll out 6 frames}} \\
    \midrule
    DINO-Foresight & 44.747 & 38.925 & 77.655 & 0.177 & 89.866 & 4.310 \\
    \addlinespace
    \multicolumn{7}{l}{\textbf{VAE (L, 16 channels)}} \\
    \quad \method (Mean) & \textbf{46.250} & \textbf{43.656} & 80.569 & 0.158 & 92.428 & 3.792 \\
    \quad \method (Best) & 44.489 & 41.817 & \textbf{82.415} & \textbf{0.150} & \textbf{92.949} & \textbf{3.712} \\
    \addlinespace
    \multicolumn{7}{l}{\textbf{PCA (16 channels)}} \\
    \quad \method (Mean) & 40.256 & 37.353 & 77.463 & 0.179 & 91.096 & 4.165 \\
    \quad \method (Best) & 39.452 & 36.528 & 80.575 & 0.164 & 91.919 & 4.023 \\
    \addlinespace
    \multicolumn{7}{l}{\textbf{PCA (1152 channels)}} \\
    \quad \method (Mean) & 41.739 & 38.986 & 78.124 & 0.194 & 89.005 & 4.491 \\
    \quad \method (Best) & 40.970 & 38.185 & 79.902 & 0.180 & 89.670 & 4.377 \\
    \addlinespace
    \multicolumn{7}{l}{\textbf{Direct  (3072 channels)}} \\
    \quad \method (Mean) & 44.479 & 41.845 & 80.092 & 0.178 & 90.345 & 4.236 \\
    \quad \method (Best) & 42.803 & 40.095 & 81.894 & 0.169 & 90.979 & 4.175 \\
    \bottomrule
    \end{tabular}%
}
\end{table*}
\begin{table*}[h!]
\centering
\caption{\textbf{Dense Forecasting Accuracy} with different diffusion spaces on Kubric. The VAE latent diffusion consistently delivers the best overall performance for dense forecasting.}
\label{tab:kubric_latent_space_ablation}

\renewcommand{\arraystretch}{0.85}

\resizebox{!}{0.475\textheight}{%
    \begin{tabular}{
    l 
    c 
    c 
    c 
    c 
    c 
    c 
    }
    \toprule
    \multirow{2}{*}{\textbf{Model}} & \multicolumn{2}{c}{\textbf{Segmentation}} & \multicolumn{2}{c}{\textbf{Depth}} & \multicolumn{2}{c}{\textbf{Normals}} \\
    \cmidrule(lr){2-3} \cmidrule(lr){4-5} \cmidrule(lr){6-7}
     & {All ($\uparrow$)} & {Mov. ($\uparrow$)} & {d1 ($\uparrow$)} & {AbsRel ($\downarrow$)} & {a3 ($\uparrow$)} & {MeanAE ($\downarrow$)} \\
    \midrule
    
    \multicolumn{7}{l}{\textit{Initial Context Length $|\mathcal{C}|=1$, roll out 11 frames}} \\
    \midrule
    DINO-Foresight & 46.729 & 5.107 & 64.370 & 0.244 & 90.618 & 2.938 \\
    \addlinespace
    \multicolumn{7}{l}{\textbf{VAE (B, 16 channels)}} \\
    \quad \method (Mean) & 48.290 & 6.611 & 68.327 & 0.205 & 93.291 & 2.144 \\
    \quad \method (Best) & \textbf{70.553} & \textbf{47.771} & \textbf{88.803} & \textbf{0.079} & 93.451 & 2.072 \\
    \addlinespace
    \multicolumn{7}{l}{\textbf{PCA (16 channels)}} \\
    \quad \method (Mean) & 49.677 & 14.478 & 68.333 & 0.278 & 88.913 & 3.303 \\
    \quad \method (Best) & 70.388 & 47.581 & 88.161 & 0.083 & 93.312 & 2.071 \\
    \addlinespace
    \multicolumn{7}{l}{\textbf{PCA (1152 channels)}} \\
    \quad \method (Mean) & 48.760 & 9.303 & 65.952 & 0.217 & 90.963 & 2.746 \\
    \quad \method (Best) & 69.488 & 45.298 & 87.380 & 0.092 & \textbf{93.820} & \textbf{1.941} \\
    \addlinespace
    \multicolumn{7}{l}{\textbf{Direct (3072 channels)}} \\
    \quad \method (Mean) & 48.476 & 10.208 & 68.492 & 0.216 & 90.727 & 2.863 \\
    \quad \method (Best) & 69.448 & 46.247 & 87.787 & 0.091 & 93.120 & 2.183 \\
    \midrule
    
    \multicolumn{7}{l}{\textit{Initial Context Length $|\mathcal{C}|=2$ , roll out 10 frames}} \\
    \midrule
    DINO-Foresight & 51.146 & 14.395 & 62.601 & 0.239 & 89.197 & 3.361 \\
    \addlinespace
    \multicolumn{7}{l}{\textbf{VAE (B, 16 channels)}} \\
    \quad \method (Mean) & 55.893 & 21.677 & 68.869 & 0.225 & \textbf{92.312} & \textbf{2.387} \\
    \quad \method (Best) & \textbf{64.840} & \textbf{37.972} & \textbf{78.058} & \textbf{0.157} & 91.659 & 2.594 \\
    \addlinespace
    \multicolumn{7}{l}{\textbf{PCA (16 channels)}} \\
    \quad \method (Mean) & 55.496 & 22.624 & 66.116 & 0.286 & 89.506 & 3.120 \\
    \quad \method (Best) & 62.645 & 34.403 & 77.716 & 0.166 & 91.145 & 2.742 \\
    \addlinespace
    \multicolumn{7}{l}{\textbf{PCA (1152 channels)}} \\
    \quad \method (Mean) & 52.017 & 14.512 & 62.255 & 0.239 & 91.174 & 2.682 \\
    \quad \method (Best) & 58.841 & 26.878 & 71.429 & 0.195 & 91.508 & 2.646 \\
    \addlinespace
    \multicolumn{7}{l}{\textbf{Direct (3072 channels)}} \\
    \quad \method (Mean) & 54.539 & 19.612 & 63.125 & 0.241 & 91.089 & 2.708 \\
    \quad \method (Best) & 60.701 & 30.480 & 78.037 & 0.165 & 91.264 & 2.713 \\
    \midrule
    
    \multicolumn{7}{l}{\textit{Initial Context Length $|\mathcal{C}|=3$, roll out 9 frames}} \\
    \midrule
    DINO-Foresight & 54.278 & 19.955 & 66.759 & 0.224 & 89.425 & 3.261 \\
    \addlinespace
    \multicolumn{7}{l}{\textbf{VAE (B, 16 channels)}} \\
    \quad \method (Mean) & 59.466 & 27.785 & 71.935 & 0.209 & \textbf{92.672} & \textbf{2.262} \\
    \quad \method (Best) & \textbf{68.254} & \textbf{43.737} & \textbf{80.092} & \textbf{0.142} & 92.314 & 2.400 \\
    \addlinespace
    \multicolumn{7}{l}{\textbf{PCA (16 channels)}} \\
    \quad \method (Mean) & 58.693 & 28.074 & 68.508 & 0.261 & 89.988 & 2.966 \\
    \quad \method (Best) & 65.935 & 39.967 & 78.727 & 0.151 & 91.796 & 2.542 \\
    \addlinespace
    \multicolumn{7}{l}{\textbf{PCA (1152 channels)}} \\
    \quad \method (Mean) & 53.959 & 18.141 & 64.738 & 0.230 & 91.198 & 2.675 \\
    \quad \method (Best) & 60.594 & 30.084 & 71.729 & 0.190 & 91.624 & 2.608 \\
    \addlinespace
    \multicolumn{7}{l}{\textbf{Direct (3072 channels)}} \\
    \quad \method (Mean) & 56.345 & 22.980 & 66.109 & 0.230 & 91.123 & 2.700 \\
    \quad \method (Best) & 62.481 & 33.867 & 79.520 & 0.154 & 91.463 & 2.672 \\
    \midrule
    
    \multicolumn{7}{l}{\textit{Initial Context Length $|\mathcal{C}|=4$, roll out 8 frames}} \\
    \midrule
    DINO-Foresight & 57.619 & 25.779 & 69.308 & 0.215 & 89.824 & 3.114 \\
    \addlinespace
    \multicolumn{7}{l}{\textbf{VAE (B, 16 channels)}} \\
    \quad \method (Mean) & 61.741 & 31.858 & 71.876 & 0.206 & \textbf{92.661} & \textbf{2.247} \\
    \quad \method (Best) & \textbf{69.864} & \textbf{46.562} & 80.498 & \textbf{0.137} & 92.622 & 2.314 \\
    \addlinespace
    \multicolumn{7}{l}{\textbf{PCA (16 channels)}} \\
    \quad \method (Mean) & 60.656 & 31.269 & 69.259 & 0.248 & 90.573 & 2.811 \\
    \quad \method (Best) & 68.039 & 43.517 & 79.437 & 0.146 & 92.186 & 2.420 \\
    \addlinespace
    \multicolumn{7}{l}{\textbf{PCA (1152 channels)}} \\
    \quad \method (Mean) & 54.820 & 19.864 & 67.186 & 0.227 & 91.222 & 2.687 \\
    \quad \method (Best) & 62.053 & 32.944 & 74.385 & 0.173 & 91.608 & 2.626 \\
    \addlinespace
    \multicolumn{7}{l}{\textbf{Direct (3072 channels)}} \\
    \quad \method (Mean) & 57.862 & 25.812 & 67.927 & 0.223 & 91.006 & 2.716 \\
    \quad \method (Best) & 64.385 & 37.248 & \textbf{80.785} & 0.147 & 91.534 & 2.638 \\
    \bottomrule
    \end{tabular}%
}
\end{table*}

\paragraph{Latent Space Spectral Analysis.}
\Cref{fig:imagenet_spectrum,fig:city_spectrum} show that uncompressed DINO features exhibit spectral characteristics similar to RGB inputs. As
Gaussian regularization (KL loss weight) on the compressed features increases, the spectrum shifts toward higher frequencies, reflecting noise injected into
the latent space. These results are consistent with recent findings in RGB domain~\cite{skorokhodov2025improvingdiffusabilityautoencoders}. Thus, the choice of VAE hyperparameters for optimal generation requires careful spectral analysis. 
\begin{figure*}[h!]
    \centering
    \input{figures/imagenet_spectrum_plot}  
    \caption{\textbf{Frequency profiles} on ImageNet 256x256. Uncompressed DINO features exhibit spectral characteristics similar to RGB inputs. As Gaussian regularization on the compressed features increases, the spectrum shifts toward higher frequencies, reflecting noise injected into the latent space.}
    \label{fig:imagenet_spectrum}
\end{figure*}
\begin{figure*}[h!]
    \centering
    \input{figures/cityscapes_spectrum_plot}  
    \caption{\textbf{Frequency profiles} on CityScapes. Uncompressed DINO features exhibit spectral characteristics similar to RGB inputs. As Gaussian regularization on the compressed features increases, the spectrum shifts toward higher frequencies, reflecting noise injected into the latent space.}
    \label{fig:city_spectrum}
\end{figure*}
\section{Qualitative Results}
\label{sec:qualitative}

\paragraph{Feature Forecasting.}%
Please refer to our \textbf{offline website} for sample animations.

\begin{figure}
    \centering
    \includegraphics[width=\linewidth]{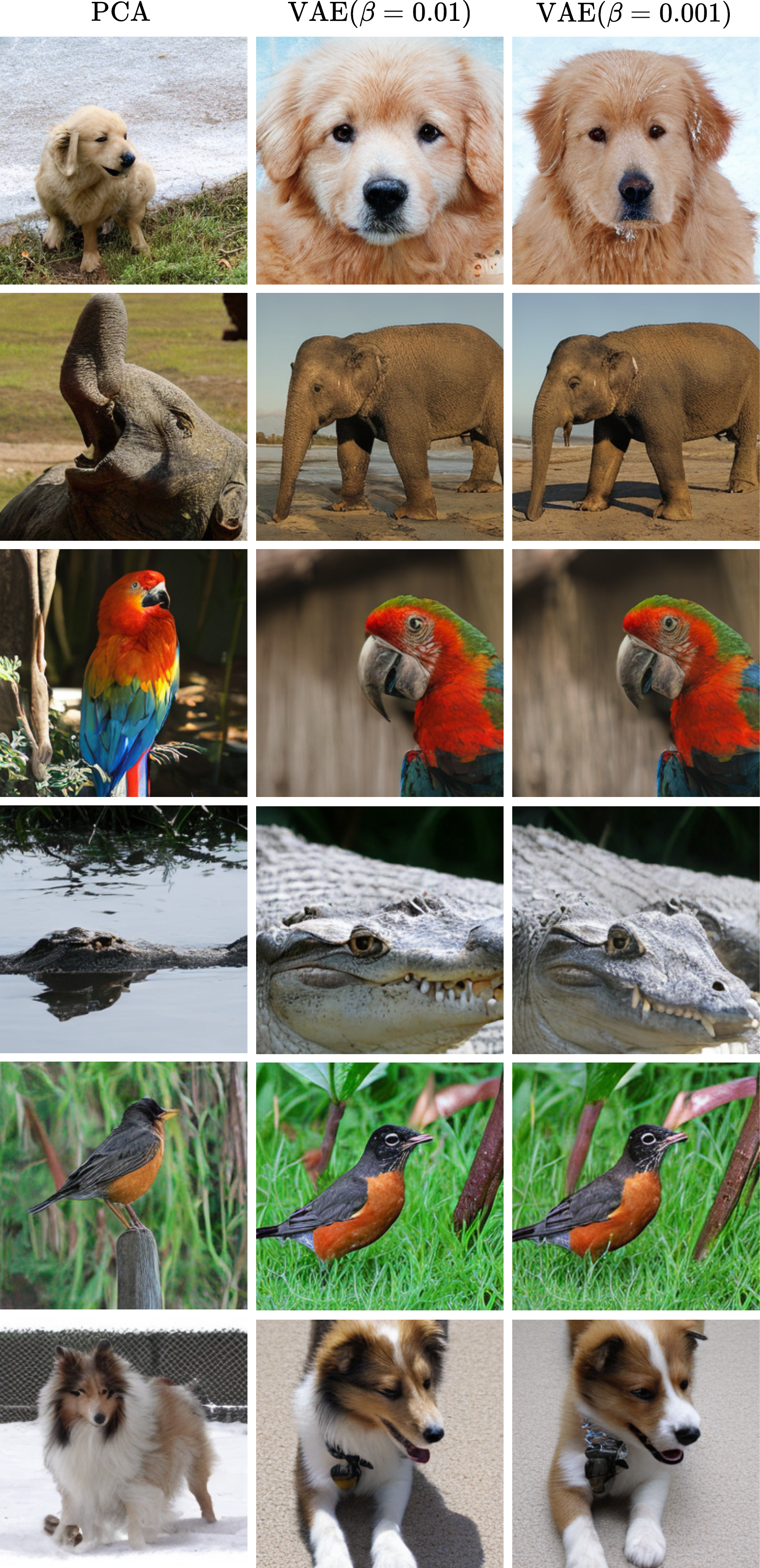}
    \caption{\textbf{Qualitative comparison of image quality} of \textbf{SiT-XL}, with ReDi guidance at 400K training steps. Diffusing VAE latents instead of PCA projections enhances fidelity, realism, and sharpness, resulting in higher quality samples.}
    \label{fig:moreimagenet}
\end{figure}

\begin{figure}
    \centering
    \includegraphics[width=\linewidth]{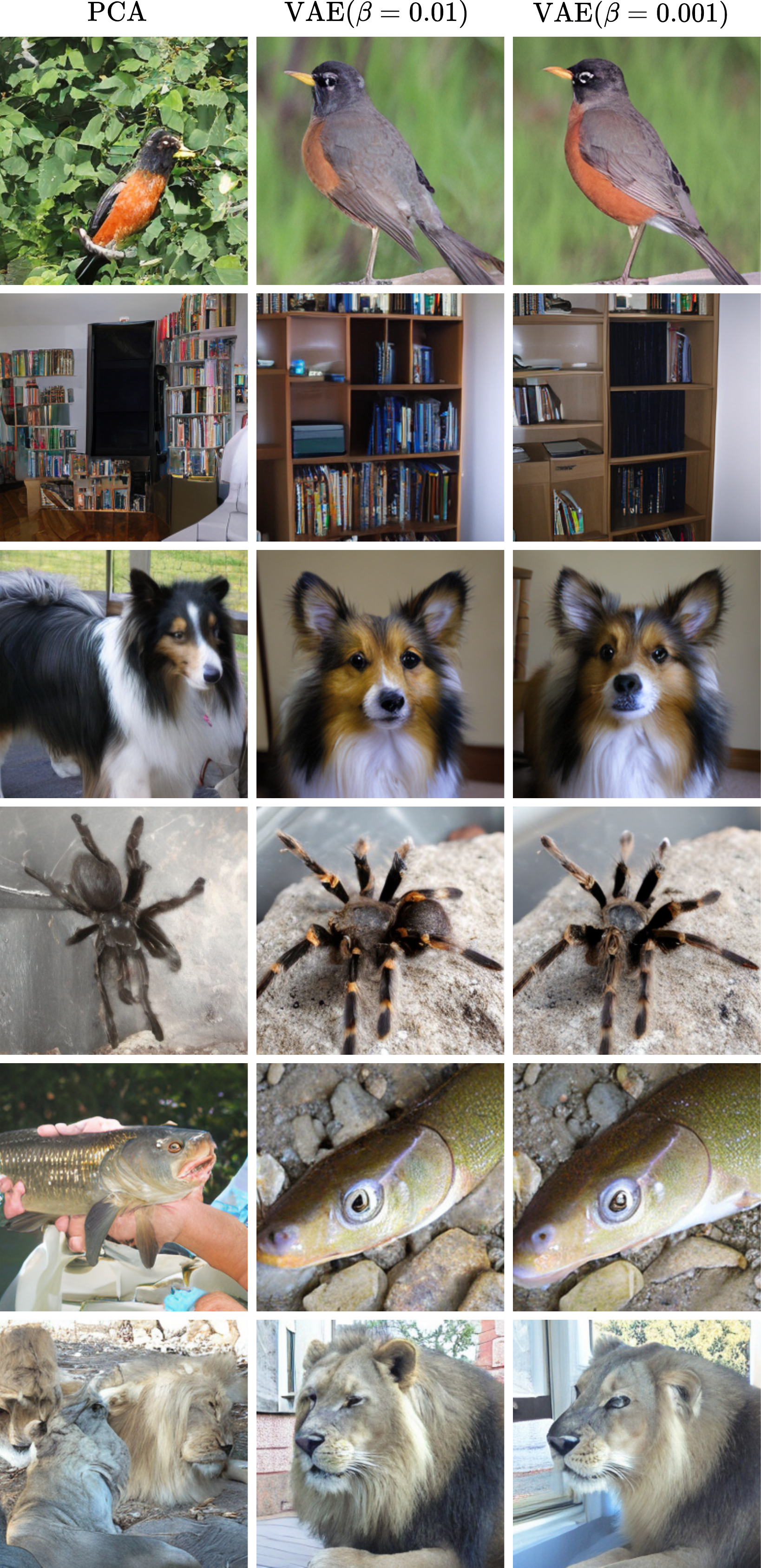}
    \caption{\textbf{Qualitative comparison of image quality} of \textbf{SiT-XL}, with ReDi guidance at 400K training steps. Diffusing VAE latents instead of PCA projections enhances fidelity, realism, and sharpness, resulting in higher quality samples.}
    \label{fig:moreimagenettwo}
\end{figure}

\begin{figure}
    \centering
    \includegraphics[width=\linewidth]{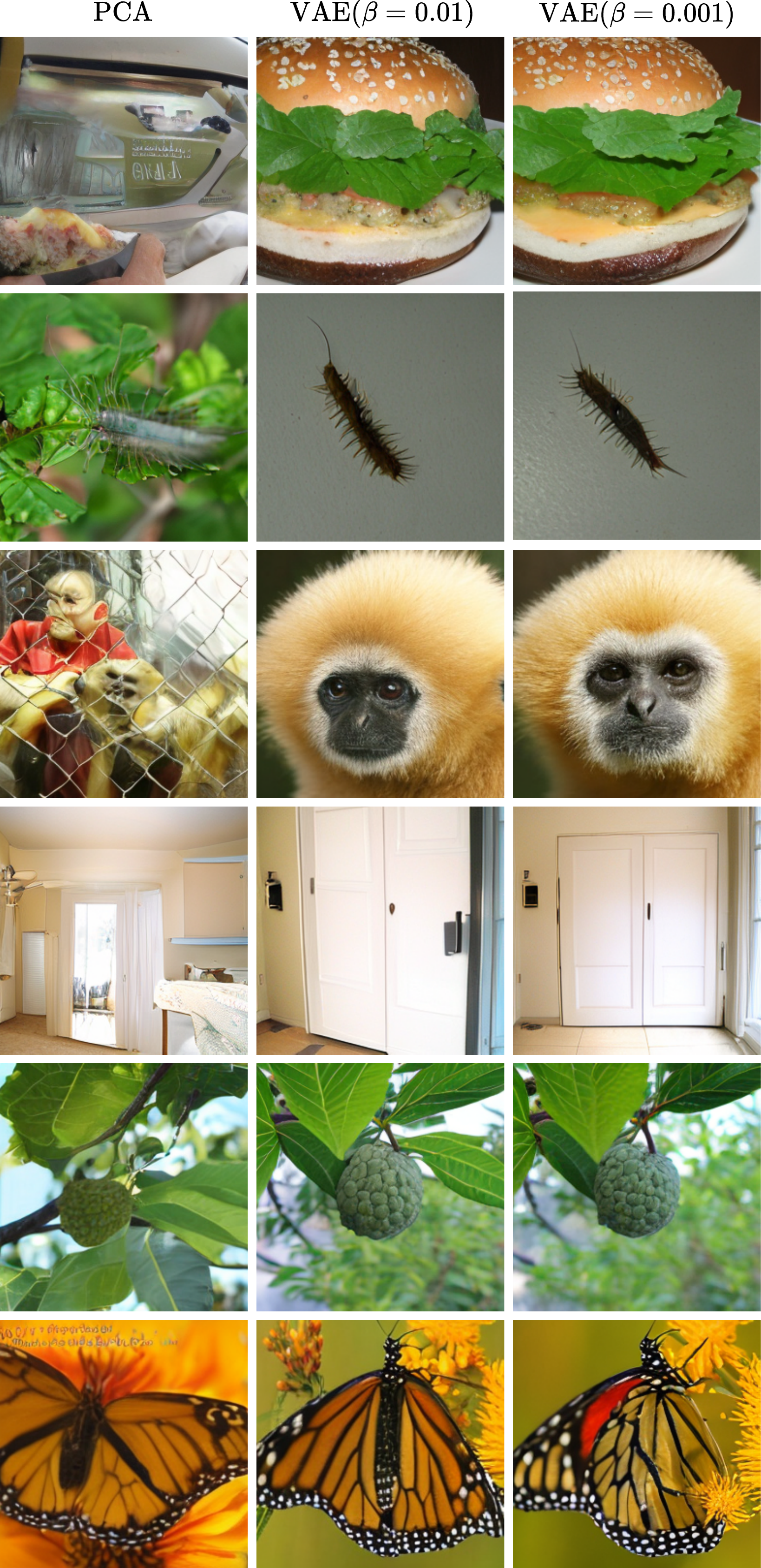}
    \caption{\textbf{Qualitative comparison of image quality} of \textbf{SiT-XL}, with ReDi guidance at 400K training steps. Diffusing VAE latents instead of PCA projections enhances fidelity, realism, and sharpness, resulting in higher quality samples.}
    \label{fig:moreimagenethree}
\end{figure}

\begin{figure}
    \centering
    \includegraphics[width=\linewidth]{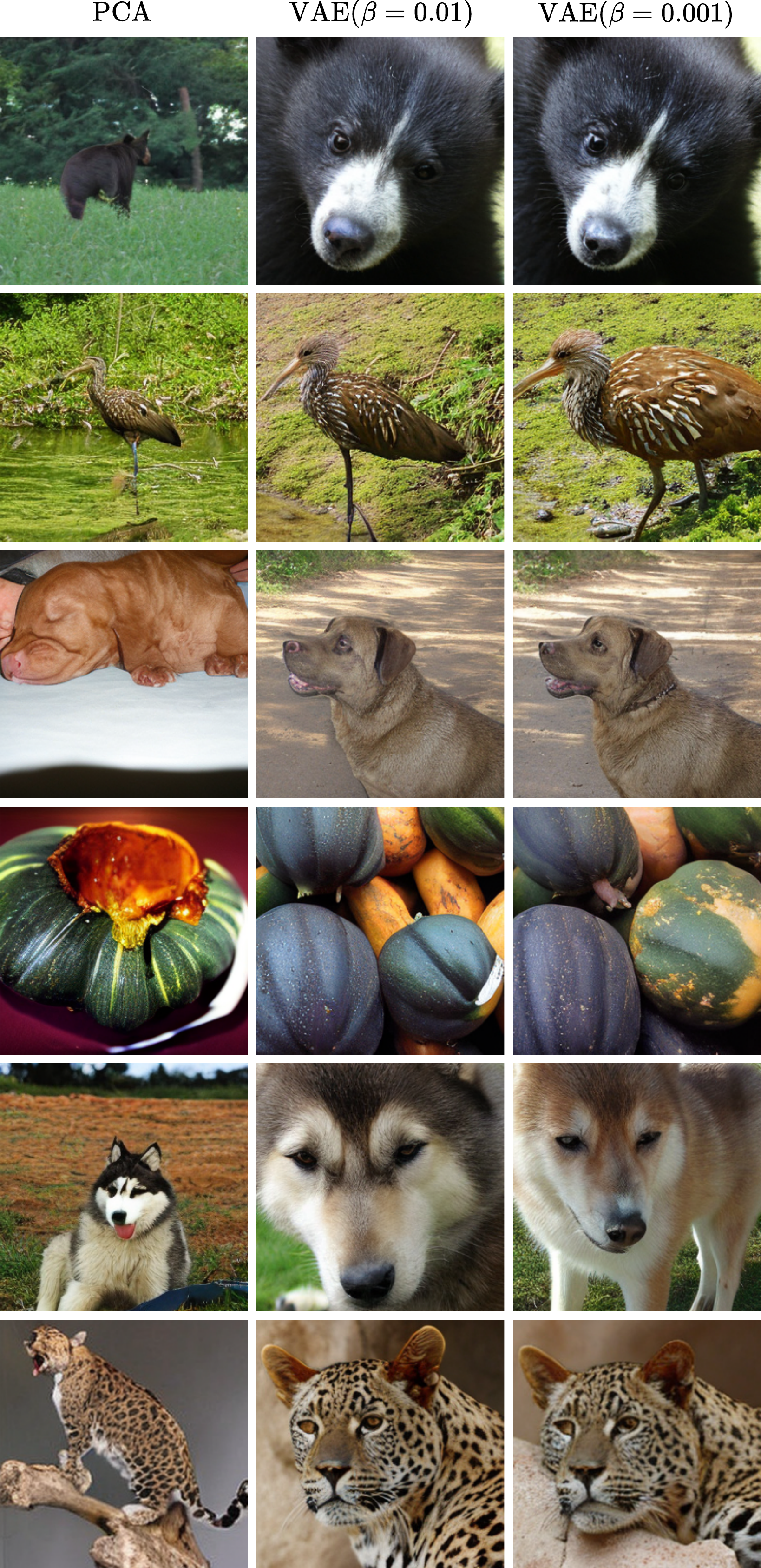}
    \caption{\textbf{Qualitative comparison of image quality} of \textbf{SiT-XL}, with ReDi guidance at 400K training steps. Diffusing VAE latents instead of PCA projections enhances fidelity, realism, and sharpness, resulting in higher quality samples.}
    \label{fig:moreimagenetfour}
\end{figure}

\paragraph{Image generation.}%
\Cref{fig:moreimagenet,fig:moreimagenettwo,fig:moreimagenethree,fig:moreimagenetfour} demonstrate that diffusing VAE latents
instead of PCA projections enhances fidelity, realism, and sharpness, yielding overall higher quality samples.


\section{Limitations and Future Work}
\label{sec:limitations}
Current limitations include (i) higher sampling latency compared to single-shot regression, (ii) mild long-horizon chroma drift, and (iii) reliance on upstream VFM domain coverage; moreover, achieving state-of-the-art video quality was not a primary objective.

Looking ahead, we plan to investigate several directions: (i) factorized video generation, i.e., training diffusion models directly in the latent space of video-centric VFMs’ VAEs~\cite{assran2025vjepa2,carreira2024scaling} paired with lightweight RGB decoders, to improve computational efficiency and long-range stability; (ii) integrating DiffusionForcing\cite{chen2025diffusion} to sustain high-fidelity predictions over extended sequences and (iii) designing a domain-specific causal diffusion architecture like \cite{wan2025}.
\clearpage
{\small\bibliographystyle{ieeenat_fullname}\bibliography{vedaldi_general,vedaldi_specific,main}}
\end{document}